\documentclass[pmlr]{jmlr} 

\newcommand{\tagdsmode}{proceedings}

\makeatletter
\newcommand{\tagdssubmission}{submission}
\newcommand{\tagdsproceedings}{proceedings}

\ifx\tagdsmode\tagdsproceedings
\else\ifx\tagdsmode\tagdssubmission
  \def\ps@jmlrtps{%
    \let\@mkboth\@gobbletwo
    \def\@oddhead{\scriptsize Under Review at the 2nd Conference on Topology, Algebra, and Geometry in Data Science\hfill}%
    \let\@evenhead\@oddhead
    \def\@oddfoot{}%
    \let\@evenfoot\@oddfoot
  }
\else
  \def\ps@jmlrtps{%
    \let\@mkboth\@gobbletwo
    \def\@oddhead{}%
    \let\@evenhead\@oddhead
    \def\@oddfoot{}%
    \let\@evenfoot\@oddfoot
  }
\fi\fi
\makeatother

\usepackage[utf8]{inputenc}

\usepackage{longtable}
\usepackage{booktabs}
\usepackage[load-configurations=version-1]{siunitx}
\usepackage{makecell}
\usepackage{xcolor}
\usepackage{array}
\usepackage{microtype}
\usepackage{xurl}   
\newcommand{\fpath}[1]{\path{#1}}

\jmlrvolume{334}
\jmlryear{2026}
\jmlrworkshop{Topology, Algebra, and Geometry in Data Science}

\title[TDL Challenge 2026]{Topological Deep Learning Challenge 2026:\\Bridging the Gap}

\author[G. Bern\'ardez, L. Telyatnikov, M. Papillon et al]{%
  \parbox{\linewidth}{\centering
    \textbf{Guillermo Bern\'ardez}$^{*}$,
    \textbf{Lev Telyatnikov}$^{*}$,
    \textbf{Mathilde Papillon}$^{*}$,
    \textbf{Marco Montagna},
    \textbf{Louisa Cornelis},
    \textbf{Louis Van Langendonck},
    \textbf{Antonio Purificato},
    \textbf{Bruno Neri},
    \textbf{David Leko},
    \textbf{Donatella Genovese},
    \textbf{Eleonora Grassucci},
    \textbf{Jan-Willem Van Looy},
    \textbf{Johan Mathe},
    \textbf{Lucia Testa},
    \textbf{Luka Beni\'c},
    \textbf{Melanie Weber},
    \textbf{Olga Zaghen},
    \textbf{Pavlo Vasylenko},
    \textbf{Raffael Theiler},
    \textbf{Scott Mahan},
    \textbf{Gaurav Khanal},
    \textbf{Valentina Sánchez},
    \textbf{Dario Loi},
    \textbf{Gabriele Onorato},
    \textbf{Prasanna Devadiga},
    \textbf{Taejin Paik},
    \textbf{Anier Velasco Sotomayor},
    \textbf{Jonas M\"uller},
    \textbf{Aarav Gupta},
    \textbf{Riyan Patel},
    \textbf{Grold Otieno Mboya},
    \textbf{Pushparaj Devadiga},
    \textbf{Adam Gohain},
    \textbf{Dominik Filipiak},
    \textbf{Michael Banf},
    \textbf{Alireza Mohammadian},
    \textbf{Amirreza Akbari},
    \textbf{Gabriel Chaviano D\'iaz},
    \textbf{Marcos Ch\'avez Mesa},
    \textbf{Ankit Grover},
    \textbf{R\'emi Bourgerie},
    \textbf{Alessio Comparini},
    \textbf{Ankit Kumar},
    \textbf{Mohammad Yusuf},
    \textbf{Devesh Kaushal},
    \textbf{Francesco Orsi},
    \textbf{Ghazal Mostofi},
    \textbf{Kolya Lettl},
    \textbf{Salwa Khattami},
    \textbf{Gabriel Franco},
    \textbf{Yelizaveta Falkouskaya},
    \textbf{Jorge Luiz Franco},
    \textbf{Kajal Singla},
    \textbf{Fernando David Espinosa I\~niguez},
    \textbf{Raj Patil},
    \textbf{Eric Dolores-Cuenca},
    \textbf{Luigi Petti},
    \textbf{Chenguang Xu},
    \textbf{Stefano Blando},
    \textbf{Christine Zeh},
    \textbf{Lennart Bastian},
    \textbf{Kaden Zheng},
    \textbf{Vihan Tiwari},
    \textbf{Sud Athreya},
    \textbf{Hisataka Kajita},
    \textbf{Hugo Micheron},
    \textbf{Rémi Devaux},
    \textbf{Antoine Jardin},
    \textbf{Tegan Emerson},
    \textbf{Olga Fink},
    \textbf{Nina Miolane}.
    \\[0.8em]
    \normalfont\normalsize
    $^{*}$\,Main organizers, equal contribution.
    Corresponding email: \texttt{topological.intelligence@gmail.com}
  }%
}

\jmlrauthors{G. Bern\'ardez, L. Telyatnikov, M. Papillon et al} 

\begin{document}

\maketitle

\begin{abstract}
This paper describes the 2026 edition of the \textit{Topological Deep Learning Challenge: Bridging the Gap}, hosted at the second Topology, Algebra, and Geometry in Data Science (TAG-DS) Conference. Where previous editions grew the modelling and data ecosystem of Topological Deep Learning (TDL), this edition set out to place Graph Neural Networks (GNNs) and Topological Neural Networks (TNNs) side by side under a single evaluation pipeline. Participants implemented recent state-of-the-art architectures in \texttt{TopoBench}, with Track 1 covering GNNs and Track 2 covering TNNs. They evaluated these implementations on shared synthetic benchmarks generated with \texttt{GraphUniverse}, which independently controls homophily, average degree and degree-distribution heaviness. The challenge yielded 68 qualifying submissions from 34 teams, spanning the graph, simplicial, cellular, hypergraph and combinatorial domains. For the first time, submissions were also tracked on a public leaderboard, which we summarize and analyze here. Our analysis finds that the differences between the submitted tracks are strongly task-dependent: the mean accuracy gap is modest on community detection, while on triangle counting the Track 1 mean error is 1.85 times the Track 2 mean. Several structurally informed GNNs also outperform the Track 2 mean on triangle counting.
\end{abstract}

\clearpage
\section{Introduction}
\label{sec:intro}

Many physical systems, from molecules to galaxies, are driven by interactions among their components: chemical interactions between atoms, functional connectivity between brain regions, or gravitational forces between stars. Conventionally, these relations are modeled with graphs, allowing Graph Neural Networks (GNNs)~\citep{scarselli2008graph} to extract key insights~\citep{kipf2017semi}. However, these systems also involve richer interactions that extend beyond the pairwise relations graphs encode. Topological Neural Networks (TNNs) have emerged to address this limitation by accounting for higher-order interactions~\citep{hajij2022topologicalDL,bodnar_2022,papillon2023architectures}, extending message passing to simplicial, cellular, hypergraph and combinatorial domains using concepts from algebraic topology~\citep{bick2023higher,barbarossa2020topological}.

Despite this shared lineage, the GNN and Topological Deep Learning (TDL) communities have largely operated in parallel. The two fields are often presented as distinct paradigms, yet they are deeply intertwined, and a growing body of work studies the interplay between standard message passing and higher-order representations. Comparisons between GNNs and TNNs remain difficult because published results typically use different codebases, datasets, tasks, and tuning budgets. This leaves a central empirical question: \emph{when does higher-order structure improve performance?}  

Building on the three previous editions of the challenge~\citep{pmlr-v221-papillon23a,bernardez2024icml,bernardez2026challenge}, the \textit{Topological Deep Learning Challenge 2026} set the stage for closing this gap. Participants were asked to implement recent state-of-the-art architectures from either literature into \texttt{TopoBench}~\citep{telyatnikov2025topobench}, the open-source platform for developing and benchmarking TDL models, and to evaluate them through a single shared pipeline on synthetic graphs generated with \texttt{GraphUniverse}~\citep{van2025graphuniverse}. Because \texttt{GraphUniverse} generates graphs with prescribed structural properties, the resulting benchmark allows us to examine how each architecture responds to changes in homophily, average degree and degree-distribution heaviness under a controlled generative design.

The challenge concluded with 68 valid submissions from 34 teams, submitted by 45 registered participants. Thirty-seven submissions implemented graph architectures in Track 1; thirty-one implemented higher-order architectures in Track 2, distributed across the simplicial, cellular, hypergraph and combinatorial domains. For the first time in the history of the challenge, submissions were additionally evaluated on a common benchmark suite and tracked on a public leaderboard. This paper details the challenge setup and submission requirements, lists all submissions and winners, and presents an analysis of the leaderboard results.

\section{Setup of the Challenge}
\label{sec:setup}

The challenge\footnote{Challenge repository: \url{https://github.com/geometric-intelligence/TopoBench}. Challenge website: \url{https://geometric-intelligence.github.io/topobench/tdl-challenge-2026/index.html}.} was hosted at the second Topology, Algebra, and Geometry in Data Science (TAG-DS) Conference and sponsored by \href{https://www.arlq.ai/}{Arlequin AI} and \href{https://www.newtheory.ai/}{New Theory}. Participants were asked to implement a recent state-of-the-art message-passing model into \texttt{TopoBench} in one of two tracks:

\begin{itemize}
\item \textbf{Track 1 --- Graph Neural Networks (GNNs).} Architectures operating on graph structures, including message-passing networks and graph transformers with structural encodings.
\item \textbf{Track 2 --- Topological Neural Networks (TNNs).} Higher-order message-passing models leveraging rich topological domains, namely simplicial, cellular, hypergraph and combinatorial complexes.
\end{itemize}

While the tracks were judged separately, the cross-track analysis uses a common evaluation protocol and controlled dataset family.

\subsection{Guidelines}
Participation was free and open to everyone. To enroll, participants had to open a Pull Request (PR) on \texttt{TopoBench}, fill out the Registration Google Form with PR and team information, and tag the PR with the appropriate track label (\texttt{track-1-gnn} or \texttt{track-2-tnn}). Teams were accepted with a maximum of two members, though larger teams were possible upon approval. A single team was permitted to submit multiple PRs, both within and across tracks. The final submission deadline was August 1st, 2026 (AoE).

Submissions were required to implement models already proposed in the literature and to cite the associated publication or pre-print in the PR description. Each PR could contain at most one core model architecture; models with several significant variants were asked to supply a separate configuration file per variant. Participants were encouraged to submit early, both to leave time to resolve integration issues and because, where multiple high-quality submissions covered the same architecture, earlier submissions were given priority.

Consistent with the aims of an open environment for sharing participation, this activity was completely voluntary. However, due to legal restrictions, individuals affiliated with institutions on the Restricted Foreign Research Institutions list were not eligible for reward outcomes, including co-authorship on this white paper.

\subsection{Submission Requirements}
\label{subsec:requirements}
A submission consisted of a Pull Request to \texttt{TopoBench} repository, titled following the format \texttt{Track: [Track1|Track2]; Team Name: <team name>; Model: <model name>}. To be valid, the PR had to pass all integration tests, including linting and formatting. Three components were required of every submission, across tracks:

\begin{enumerate}
\item \textbf{Backbone.} The model implemented as a \texttt{torch.nn.Module} and placed in \fpath{topobench/nn/backbones/{domain}/{model_name}.py}, where \texttt{\{domain\}} is one of \texttt{graph}, \texttt{simplicial}, \texttt{cell}, \texttt{hypergraph}, \texttt{combinatorial} or \texttt{non\_relational}. 
\item \textbf{Hydra configuration.} A YAML file at \fpath{configs/model/{domain}/{model_name}.yaml} specifying the full \texttt{TBModel} composition: feature encoder, backbone, backbone wrapper and readout.
\item \textbf{Results file.} The \texttt{results.json} produced by running the provided evaluation notebook, \fpath{TopoBench/2026_tdl_challenge/run_evaluation.ipynb}, on the submitted model. This file records the model's performance on the benchmark tasks and available resource measurements, and is what feeds the public leaderboard.
\end{enumerate}

Beyond the backbone, participants were encouraged to exploit \texttt{TopoBench}'s modular architecture by contributing custom feature encoders, readouts, backbone wrappers or losses where their chosen model required them, each inheriting from the corresponding abstract base class and discovered automatically by the framework. This was a deliberate design choice: it enriches the ecosystem with reusable components rather than with monolithic model files.

\paragraph{Testing.} All contributed files had to pass the pre-existing unit tests, and any method or class not already covered had to be accompanied by new tests mirroring the source directory structure. Coverage, measured with Codecov, had to match or exceed 93\%. Each submission also had to fill in \fpath{test/pipeline/test_pipeline.py} to demonstrate that the full training pipeline runs end-to-end, and the evaluation notebook had to run successfully with the submitted model.

\subsection{Evaluation Method}
\label{subsec:evaluation}
To decide the winners in each track, a panel of \texttt{TopoBench} maintainers and collaborators voted on their preferred submissions, and the Condorcet method was used to produce a ranking. Reviewers were asked to base their rankings on the following criteria, which were visible to participants throughout:

\begin{itemize}
\item \textbf{Correctness:} Does the submission correctly implement the model as described in the literature? Modifications required to respect \texttt{TopoBench}'s computational constraints were permitted.
\item \textbf{Code quality:} How readable and clean is the implementation? How well does the submission respect the requirements, including unit tests and memory usage?
\item \textbf{Benchmark on GraphUniverse datasets:} Is the model correctly benchmarked on the provided datasets? Does it produce reasonable results?
\item \textbf{Documentation \& tests:} Do docstrings clearly describe the code, and do they make explicit reference to the original equations in the associated paper and/or the original implementation? Are unit tests robust?
\end{itemize}

The award criteria assess the quality and reusability of model implementations, not their final benchmark performance. No minimum training performance was required, and the highest leaderboard score did not determine the winner. The leaderboard supports the comparative analysis in Section~\ref{sec:results}.

For the first time, reviewers were permitted to use AI tools to help manage the volume of submissions. In Track 1, an AI reviewer agent was given the same instructions as the human reviewers. It graded each submission against the rubric and flagged the bottom tier of submissions. Track 1 reviewers could use these flags at their discretion. Reviewers were asked to disclose their use of AI in their responses.

\subsection{Prizes}
With thanks to our sponsors, this edition offered cash prizes of \$1{,}000 USD for first place and \$400 USD for second place in each track, with Track 1 sponsored by \href{https://www.newtheory.ai/}{New Theory} and Track 2 by \href{https://www.arlq.ai/}{Arlequin AI}. This year, honorable mentions also carried cash prizes, with \$700 USD split across outstanding submissions. In addition, selected teams were invited for research internships at the Geometric Intelligence Lab (University of California, Santa Barbara) and the Intelligent Maintenance and Operations Systems (IMOS) Lab (\'Ecole Polytechnique F\'ed\'erale de Lausanne, Switzerland).

\subsection{Software Practices}
All submitted code had to comply with the challenge's GitHub Action workflow, passing all tests, linting and formatting checks. To ensure consistency, participants working in the simplicial, cellular or combinatorial domains were asked to use TopoNetX's~\citep{hajij2024topox} classes to manage the corresponding complexes.

\section{Benchmarking Protocol}
\label{sec:protocol}

The defining novelty of this edition is the use of a shared evaluation pipeline and dataset family for the cross-track comparison. This section describes that protocol.

\subsection{Controlled Synthetic Data}
Evaluation datasets were generated with \texttt{GraphUniverse}~\citep{van2025graphuniverse}, a framework for sampling synthetic graphs with prescribed structural properties. Three structural axes were varied independently:

\begin{itemize}
\item \textbf{Homophily}, at three levels: low ($0$--$0.1$), mid ($0.4$--$0.6$) and high ($0.9$--$1.0$).
\item \textbf{Average degree}, at two levels: low ($1$--$2.5$) and high ($4$--$5$).
\item \textbf{Power-law exponent $\gamma$} of the degree distribution, at two levels: low ($1.5$--$2$, heavy-tailed) and high ($4$--$5$, homogeneous).
\end{itemize}

The full grid therefore contains $3 \times 2 \times 2 = 12$ configurations. The prescribed evaluation uses three training seeds per configuration, totaling 36 runs per task and 72 per model variant. This controlled design allows us to examine how model performance changes across prescribed structural regimes.

\subsection{Tasks and Metrics}
We consider two tasks, probing community structure and local substructure:

\begin{itemize}
\item \textbf{Community detection}, a node-classification task scored by accuracy (higher is better). This task is well aligned with the inductive bias of standard message passing: it rewards models that aggregate and smooth information over neighbourhoods.
\item \textbf{Triangle counting}, a regression task scored by $\mathrm{MSE}_{\mathrm{test}}/\sum_{G\in\mathcal D_{\mathrm{test}}}\tau(G)$ (lower is better), where $\tau(G)$ counts triangles in each test graph's undirected edge structure after preprocessing. Counting triangles is provably beyond the reach of $1$-WL-bounded message-passing networks.
\end{itemize}

\subsection{Leaderboard}
Results were published on a public leaderboard\footnote{\url{https://geometric-intelligence.github.io/topobench/leaderboard/index.html}}. For each structural setting, scores are reported as the mean $\pm$ standard deviation of the in-distribution test metric across three training seeds; summaries across settings include up to 36 runs per task. The analysis uses the submitted result reports, with one variant per PR selected as described in Section~\ref{sec:results}.

Two caveats underscore these results. First, the scores describe the implementations and configurations submitted by participants rather than the best achievable performance of each method. Second, implementations were contributed by different participants with differing familiarity with the underlying papers. While the review process checked for correctness as part of the rubric, residual implementation variance potentially impacts leaderboard performance.

\section{Submissions and Winners}
\label{sec:submissions}

The challenge received 68 valid submissions from 34 teams, submitted by 45 registered participants. Participation was close to evenly split between the two tracks: 37 submissions in Track 1 and 31 in Track 2. Within Track 2, submissions covered all four higher-order domains supported by \texttt{TopoBench}: ten simplicial, six cellular, seven hypergraph and eight combinatorial. Table~\ref{tab:submissions_overview} lists all submitted models, grouped by track and domain, with references to the underlying methods and links to their implementations. Additionally, Table~\ref{tab:teams} in Appendix~\ref{app:teams} lists the co-authors belonging to each team, as well as the awards each team received.

\clearpage
\begingroup
\begin{table}[!ht]
\centering
\definecolor{cataloguePlum}{HTML}{7D0055}
\definecolor{catalogueGold}{HTML}{FAC710}
\definecolor{catalogueGrey}{HTML}{808080}
\definecolor{catalogueTeal}{HTML}{C1E8E6}
\fontsize{9}{10.2}\selectfont
\setlength{\tabcolsep}{0pt}
\renewcommand{\arraystretch}{1.0}
\newcommand{\cataloguegroup}[3]{%
  \par\noindent\colorbox{#1}{%
    \parbox{\dimexpr\linewidth-2\fboxsep\relax}{%
      \strut\textbf{#2}\hfill #3 submissions\strut}}\par\smallskip}
\begin{minipage}[t]{0.485\textwidth}
\textbf{Track 1 --- GNNs}\par\smallskip
\cataloguegroup{catalogueGrey!18}{Graph}{37}
\begin{tabular}{@{}>{\raggedright\arraybackslash}p{\dimexpr\linewidth-24pt\relax}@{\hspace{5pt}}>{\raggedleft\arraybackslash}p{19pt}@{}}
\textit{Model} & \textit{PR} \\[2pt]
AdvDIFFormer~\citep{tdl26_wu2025superchargingdiffusion} & \href{https://github.com/geometric-intelligence/TopoBench/pull/399}{399} \\
BuNN~\citep{tdl26_bamberger2025bundlegraphs} & \href{https://github.com/geometric-intelligence/TopoBench/pull/319}{319} \\
CoGNN~\citep{tdl26_finkelshtein2024cooperativenetworks} & \href{https://github.com/geometric-intelligence/TopoBench/pull/369}{369} \\
Conn-NSD~\citep{tdl26_barbero2022sheaflaplacians} & \href{https://github.com/geometric-intelligence/TopoBench/pull/327}{327} \\
DNSD~\citep{tdl26_bourgerie2026deepdiffusion} & \href{https://github.com/geometric-intelligence/TopoBench/pull/375}{375} \\
DSNN~\citep{tdl26_fiorini2026sheavesawakening} & \href{https://github.com/geometric-intelligence/TopoBench/pull/412}{412} \\
ESC-GNN~\citep{tdl26_yan2024efficientpower} & \href{https://github.com/geometric-intelligence/TopoBench/pull/419}{419} \\
FAGCN~\citep{tdl26_bo2021beyondnetworks} & \href{https://github.com/geometric-intelligence/TopoBench/pull/382}{382} \\
FAGCN~\citep{tdl26_bo2021beyondnetworks} & \href{https://github.com/geometric-intelligence/TopoBench/pull/388}{388} \\
FiGURe~\citep{tdl26_ekbote2023figureaugmentations} & \href{https://github.com/geometric-intelligence/TopoBench/pull/360}{360} \\
GAMMA~\citep{tdl26_ghazizadeh2025gammagnns} & \href{https://github.com/geometric-intelligence/TopoBench/pull/411}{411} \\
GATE~\citep{tdl26_mustafa2024gateneighbors} & \href{https://github.com/geometric-intelligence/TopoBench/pull/355}{355} \\
GAUGE~\citep{tdl26_sun2026arebundles} & \href{https://github.com/geometric-intelligence/TopoBench/pull/396}{396} \\
GFT~\citep{tdl26_wang2024gftvocabulary} & \href{https://github.com/geometric-intelligence/TopoBench/pull/334}{334} \\
GPR-GNN~\citep{tdl26_chien2021adaptivenetwork} & \href{https://github.com/geometric-intelligence/TopoBench/pull/352}{352} \\
GREAD~\citep{tdl26_choi2023greadnetworks} & \href{https://github.com/geometric-intelligence/TopoBench/pull/415}{415} \\
GREAD~\citep{tdl26_choi2023greadnetworks} & \href{https://github.com/geometric-intelligence/TopoBench/pull/420}{420} \\
GRIT~\citep{tdl26_ma2023graphpassing} & \href{https://github.com/geometric-intelligence/TopoBench/pull/370}{370} \\
GSN~\citep{tdl26_bouritsas2023improvingcounting} & \href{https://github.com/geometric-intelligence/TopoBench/pull/407}{407} \\
HeroFilter~\citep{tdl26_zhang2025herofilterrelations} & \href{https://github.com/geometric-intelligence/TopoBench/pull/384}{384} \\
HiD-Net~\citep{tdl26_li2024generalizedgraphs} & \href{https://github.com/geometric-intelligence/TopoBench/pull/335}{335} \\
HOD-GNN~\citep{tdl26_eitan2026expressivederivatives} & \href{https://github.com/geometric-intelligence/TopoBench/pull/402}{402} \\
HyMN~\citep{tdl26_southern2025balancingcentrality} & \href{https://github.com/geometric-intelligence/TopoBench/pull/421}{421} \\
ISN~\citep{tdl26_caralt2026necessitylaplacians} & \href{https://github.com/geometric-intelligence/TopoBench/pull/374}{374} \\
JacobiConv~\citep{tdl26_wang2022hownetworks} & \href{https://github.com/geometric-intelligence/TopoBench/pull/383}{383} \\
LINKX~\citep{tdl26_lim2021largemethods} & \href{https://github.com/geometric-intelligence/TopoBench/pull/361}{361} \\
MTGCN~\citep{tdl26_pei2024multimixing} & \href{https://github.com/geometric-intelligence/TopoBench/pull/400}{400} \\
NSD~\citep{tdl26_bodnar2022neuralgnns} & \href{https://github.com/geometric-intelligence/TopoBench/pull/332}{332} \\
NSP~\citep{tdl26_suk2022surfingsheaf} & \href{https://github.com/geometric-intelligence/TopoBench/pull/326}{326} \\
NSP + Holonomy~\citep{tdl26_suk2022surfingsheaf} & \href{https://github.com/geometric-intelligence/TopoBench/pull/373}{373} \\
OGFormer~\citep{tdl26_zhang2025graphclassification} & \href{https://github.com/geometric-intelligence/TopoBench/pull/378}{378} \\
PolyFilter GNN (Favard)~\citep{tdl26_guo2023graphbases} & \href{https://github.com/geometric-intelligence/TopoBench/pull/354}{354} \\
Polynormer~\citep{tdl26_deng2024polynormertime} & \href{https://github.com/geometric-intelligence/TopoBench/pull/345}{345} \\
r-lMPNN~\citep{tdl26_paolino2024weisfeilerlearning} & \href{https://github.com/geometric-intelligence/TopoBench/pull/404}{404} \\
SGFormer~\citep{tdl26_wu2023sgformerrepresentations} & \href{https://github.com/geometric-intelligence/TopoBench/pull/359}{359} \\
SheafAN~\citep{tdl26_barbero2022sheafnetworks} & \href{https://github.com/geometric-intelligence/TopoBench/pull/322}{322} \\
TopoPolynormer~\citep{tdl26_deng2024polynormertime} & \href{https://github.com/geometric-intelligence/TopoBench/pull/353}{353} \\
\end{tabular}
\end{minipage}\hfill
\begin{minipage}[t]{0.485\textwidth}
\textbf{Track 2 --- TNNs}\par\smallskip
\cataloguegroup{cataloguePlum!12}{Simplicial}{10}
\begin{tabular}{@{}>{\raggedright\arraybackslash}p{\dimexpr\linewidth-24pt\relax}@{\hspace{5pt}}>{\raggedleft\arraybackslash}p{19pt}@{}}
\textit{Model} & \textit{PR} \\[2pt]
Bi-SCNN~\citep{tdl26_yan2025binarizednetworks} & \href{https://github.com/geometric-intelligence/TopoBench/pull/387}{387} \\
COSIMO~\citep{tdl26_einizade2025continuousnetworks} & \href{https://github.com/geometric-intelligence/TopoBench/pull/392}{392} \\
CTNN~\citep{tdl26_hajij2025copresheafframework} & \href{https://github.com/geometric-intelligence/TopoBench/pull/405}{405} \\
Dir-SNN~\citep{tdl26_lecha2025highernetworks} & \href{https://github.com/geometric-intelligence/TopoBench/pull/328}{328} \\
DirSNN (memory-safe, harmonic-aware)~\citep{tdl26_lecha2025highernetworks} & \href{https://github.com/geometric-intelligence/TopoBench/pull/350}{350} \\
GSAN~\citep{tdl26_battiloro2024generalizednetworks} & \href{https://github.com/geometric-intelligence/TopoBench/pull/365}{365} \\
HiGCN~\citep{tdl26_huang2024highercomplexes} & \href{https://github.com/geometric-intelligence/TopoBench/pull/364}{364} \\
MPSN~\citep{tdl26_bodnar2021weisfeilernetworks} & \href{https://github.com/geometric-intelligence/TopoBench/pull/376}{376} \\
MPSN~\citep{tdl26_bodnar2021weisfeilernetworks} & \href{https://github.com/geometric-intelligence/TopoBench/pull/393}{393} \\
PolynomialFilterTNN (SNN)~\citep{tdl26_ebli2020simplicialnetworks} & \href{https://github.com/geometric-intelligence/TopoBench/pull/414}{414} \\
\end{tabular}
\par\medskip
\cataloguegroup{catalogueGold!25}{Cellular}{6}
\begin{tabular}{@{}>{\raggedright\arraybackslash}p{\dimexpr\linewidth-24pt\relax}@{\hspace{5pt}}>{\raggedleft\arraybackslash}p{19pt}@{}}
AirTNN~\citep{tdl26_fiorellino2024topologicalair} & \href{https://github.com/geometric-intelligence/TopoBench/pull/329}{329} \\
AirTNN~\citep{tdl26_fiorellino2024topologicalair} & \href{https://github.com/geometric-intelligence/TopoBench/pull/394}{394} \\
Cellular Transformer~\citep{tdl26_ballester2024attendingtransformer} & \href{https://github.com/geometric-intelligence/TopoBench/pull/371}{371} \\
CIN~\citep{tdl26_bodnar2021weisfeilernetworksa} & \href{https://github.com/geometric-intelligence/TopoBench/pull/363}{363} \\
DiffLift-SMCN~\citep{tdl26_franco2026differentiablenetworks,tdl26_eitan2025topologicalexpressivity} & \href{https://github.com/geometric-intelligence/TopoBench/pull/406}{406} \\
SheafTSP~\citep{tdl26_tandon2026consistentsheaves} & \href{https://github.com/geometric-intelligence/TopoBench/pull/410}{410} \\
\end{tabular}
\par\medskip
\cataloguegroup{catalogueTeal!70}{Hypergraph}{7}
\begin{tabular}{@{}>{\raggedright\arraybackslash}p{\dimexpr\linewidth-24pt\relax}@{\hspace{5pt}}>{\raggedleft\arraybackslash}p{19pt}@{}}
DPHGNN~\citep{tdl26_saxena2024dphgnnnetworks} & \href{https://github.com/geometric-intelligence/TopoBench/pull/398}{398} \\
DSHN~\citep{tdl26_mule2026directionalhypergraphs} & \href{https://github.com/geometric-intelligence/TopoBench/pull/408}{408} \\
EHNN~\citep{tdl26_kim2022equivariantnetworks} & \href{https://github.com/geometric-intelligence/TopoBench/pull/366}{366} \\
HyperGCN~\citep{tdl26_yadati2019hypergcnhypergraphs} & \href{https://github.com/geometric-intelligence/TopoBench/pull/413}{413} \\
PhenomNN~\citep{tdl26_wang2023fromnetworks} & \href{https://github.com/geometric-intelligence/TopoBench/pull/367}{367} \\
SheafHyperGNN~\citep{tdl26_duta2023sheafnetworks} & \href{https://github.com/geometric-intelligence/TopoBench/pull/321}{321} \\
WHNN~\citep{tdl26_duta2025wassersteinnetwork} & \href{https://github.com/geometric-intelligence/TopoBench/pull/417}{417} \\
\end{tabular}
\par\medskip
\cataloguegroup{cataloguePlum!6!catalogueTeal!35}{Combinatorial}{8}
\begin{tabular}{@{}>{\raggedright\arraybackslash}p{\dimexpr\linewidth-24pt\relax}@{\hspace{5pt}}>{\raggedleft\arraybackslash}p{19pt}@{}}
CTNN (copresheaf, gated)~\citep{tdl26_hajij2025copresheafframework} & \href{https://github.com/geometric-intelligence/TopoBench/pull/418}{418} \\
ETNN~\citep{tdl26_battiloro2025enetworks} & \href{https://github.com/geometric-intelligence/TopoBench/pull/320}{320} \\
ETNN-CoordinatePolicy (structural-LapPE)~\citep{tdl26_battiloro2025enetworks} & \href{https://github.com/geometric-intelligence/TopoBench/pull/391}{391} \\
ETNN-LapPE~\citep{tdl26_battiloro2025enetworks} & \href{https://github.com/geometric-intelligence/TopoBench/pull/362}{362} \\
HMC~\citep{hajij2022topologicalDL} & \href{https://github.com/geometric-intelligence/TopoBench/pull/416}{416} \\
SMCN~\citep{tdl26_eitan2025topologicalexpressivity} & \href{https://github.com/geometric-intelligence/TopoBench/pull/397}{397} \\
TopNets~\citep{tdl26_verma2024topologicalcontinuous} & \href{https://github.com/geometric-intelligence/TopoBench/pull/333}{333} \\
TopoU-Net~\citep{tdl26_gaurav2026topoudomains} & \href{https://github.com/geometric-intelligence/TopoBench/pull/372}{372} \\
\end{tabular}
\end{minipage}
\caption{All 68 submissions, grouped by track and domain, with models ordered alphabetically within each domain. Citations identify the underlying methods, and PR numbers link to each implementation. Repeated models denote separate submissions. Appendix~\ref{app:teams} lists teams; Appendix~\ref{app:leaderboards} reports scores for the 65 entries in the comparative analysis.}
\label{tab:submissions_overview}
\end{table}

\clearpage
\endgroup

\subsection{Winners and Honorable Mentions}

The winners were announced publicly at the second TAG-DS Conference, on social media, and on the official challenge website. We emphasize that regardless of final rankings, the organizers and reviewers were impressed by the quality of the submissions and their contribution to the \texttt{TopoBench} ecosystem. We warmly congratulate all participants.

\paragraph{Track 1 --- Graph Neural Networks.}
\begin{itemize}
\item \textbf{First place (\$1{,}000):} \textit{DSNN}, Team \textit{howyadoin} (Dario Loi and Gabriele Onorato; PR \#412).
\item \textbf{Joint second place (\$200 each):} \textit{GRIT}, Team \textit{TripleA} (Amirreza Akbari and Alireza Mohammadian; PR \#370), and \textit{HOD-GNN}, Team \textit{SweetLesson} (Aarav Gupta and Riyan Patel; PR \#402).
\end{itemize}

\paragraph{Track 2 --- Topological Neural Networks.}
The Condorcet evaluation resulted in a tie, and the Track 2 prize was divided equally between the two co-winners:
\begin{itemize}
\item \textbf{Joint first place (\$700 each):} \textit{ETNN-CoordinatePolicy}, Team \textit{E(n)igma} (Gaurav Khanal; PR \#391), and \textit{TopoU-Net}, Team \textit{TripleA} (Amirreza Akbari and Alireza Mohammadian; PR \#372).
\end{itemize}

\paragraph{Honorable mentions.}
\begin{itemize}
\item \textbf{Excellent portfolio (\$150 each):} Teams \textit{howyadoin} (Dario Loi and Gabriele Onorato) and \textit{TripleA} (Amirreza Akbari and Alireza Mohammadian), recognised for multiple high-quality contributions across both tracks.
\item \textbf{Novel infrastructure with differentiable liftings (\$200):} \textit{DiffLift-SMCN}, Team \textit{One Ring to Lift Them All} (Jorge Luiz Franco; PR \#406).
\item \textbf{Outstanding supplementary analysis (\$200):} \textit{Neural Sheaf Propagation + Holonomy Measurement}, Team \textit{Sheafu} (Ankit Grover and R\'emi Bourgerie; PR \#373 with a companion paper \citet{grover2026sheafneuralnetworksuse}).
\end{itemize}

\section{Leaderboard Results and Analysis}
\label{sec:results}


This section presents the first cross-track comparison in the history of the challenge. We include 65 entries\footnote{The main analysis excludes PRs \#319 and \#327 because their reported test-graph triangle totals differ from those of the reference cohort, and PR \#382 because it covers all 12 structural settings with only one training seed.}: 34 in Track 1 and 31 in Track 2. For each PR, we consider variants of the submitted model with complete benchmark results, select the variant with the highest mean in-distribution community-detection accuracy over the full evaluation grid, and use that same variant for both tasks\footnote{For PR \#391, this is the structural-LapPE variant.}. For each task, we first average the three seed scores within each setting, then average the 12 setting means to obtain one score per PR. Track and domain means give each PR equal weight. Table~\ref{tab:agg_track} reports the per-track aggregates, and Table~\ref{tab:agg_domain} breaks them down by topological domain. Complete leaderboards for both tasks are given in Appendix~\ref{app:leaderboards}.

\begin{table}[htbp]
\centering
\small
\begin{tabular}{l r r r r r r r}
\toprule
& & & \multicolumn{3}{c}{\textbf{Homophily}} & \multicolumn{2}{c}{\textbf{Avg.\ degree}} \\
\cmidrule(lr){4-6}\cmidrule(lr){7-8}
\textbf{Track} & \textbf{$n$} & \textbf{All} & \textbf{Low} & \textbf{Mid} & \textbf{High} & \textbf{Low} & \textbf{High} \\
\midrule
\multicolumn{8}{l}{\emph{Community detection} --- accuracy ($\uparrow$)} \\
\quad Track 1 (GNN) & 34 & 0.4241 & 0.3088 & 0.3869 & 0.5765 & 0.3991 & 0.4491 \\
\quad Track 2 (TNN) & 31 & 0.4343 & 0.3061 & 0.4000 & 0.5970 & 0.4099 & 0.4588 \\
\midrule
\multicolumn{8}{l}{\emph{Triangle counting} --- MSE per triangle ($\downarrow$)} \\
\quad Track 1 (GNN) & 34 & 0.7097 & 0.0887 & 0.6099 & 1.4304 & 0.3579 & 1.0614 \\
\quad Track 2 (TNN) & 31 & 0.3828 & 0.0428 & 0.3251 & 0.7805 & 0.1858 & 0.5798 \\
\bottomrule
\end{tabular}
\caption{Mean in-distribution test score per track, averaged over submissions in the 65-entry analysis cohort (Section~\ref{sec:results}). Columns marginalise the GraphUniverse configuration grid over homophily and average degree. PRs \#319, \#327 and \#382 are excluded.}
\label{tab:agg_track}
\end{table}

\begin{figure}[htbp]
\centering
\includegraphics[width=\textwidth]{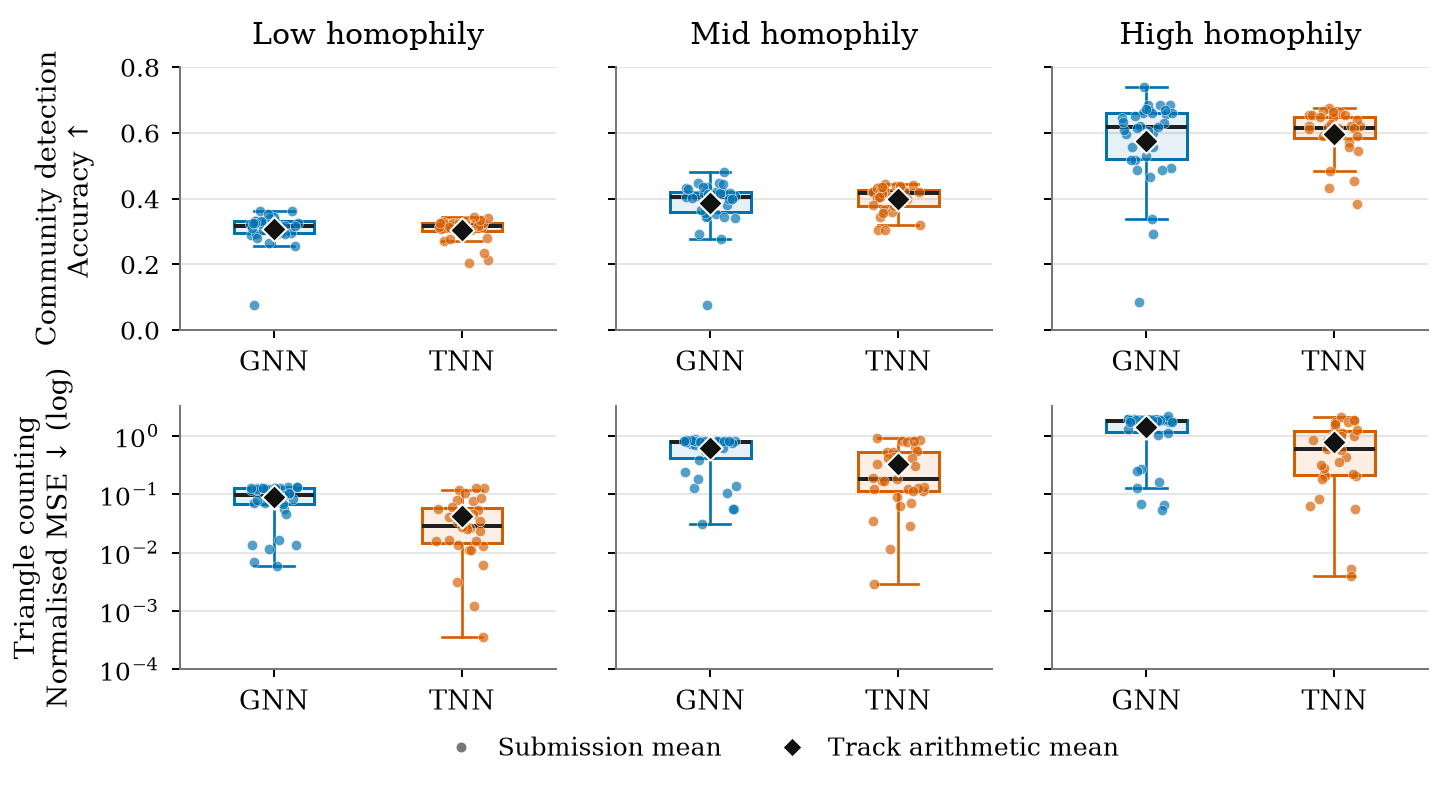}
\caption{Submission-score distributions by homophily, complementing Table~\ref{tab:agg_track}. Each dot is one submission's mean over four degree--power-law settings and three seeds per setting (34 GNNs; 31 TNNs). Boxes span the interquartile range, with a line marking the median. Whiskers extend to the most extreme observations within 1.5 interquartile ranges below the first quartile or above the third quartile. Diamonds mark arithmetic track means. Axes are shared within each task; triangle-counting error is shown on a logarithmic scale.}
\label{fig:score_distributions_homophily}
\end{figure}

\begin{table}[htbp]
\centering
\small
\begin{tabular}{l r r r r r r r r r}
\toprule
& & \multicolumn{4}{c}{\textbf{Community detection} ($\uparrow$)} & \multicolumn{4}{c}{\textbf{Triangle counting} ($\downarrow$)} \\
\cmidrule(lr){3-6}\cmidrule(lr){7-10}
\textbf{Domain} & \textbf{$n$} & \textbf{All} & \textbf{$h_{lo}$} & \textbf{$h_{mid}$} & \textbf{$h_{hi}$} & \textbf{All} & \textbf{$h_{lo}$} & \textbf{$h_{mid}$} & \textbf{$h_{hi}$} \\
\midrule
Graph (T1) & 34 & 0.4241 & 0.3088 & 0.3869 & 0.5765 & 0.7097 & 0.0887 & 0.6099 & 1.4304 \\
Simplicial & 10 & 0.4411 & 0.3141 & 0.4030 & 0.6062 & 0.2586 & 0.0119 & 0.2184 & 0.5455 \\
Cell & 6 & 0.4356 & 0.2846 & 0.3994 & 0.6227 & 0.3144 & 0.0467 & 0.2652 & 0.6314 \\
Hypergraph & 7 & 0.4451 & 0.3196 & 0.4067 & 0.6090 & 0.6505 & 0.0867 & 0.5253 & 1.3393 \\
Combinatorial & 8 & 0.4155 & 0.3004 & 0.3907 & 0.5555 & 0.3551 & 0.0403 & 0.3281 & 0.6970 \\
\bottomrule
\end{tabular}
\caption{Mean in-distribution test score per topological domain, broken down by homophily regime. Track 1 submissions all operate on the graph domain. The 65-entry analysis cohort is defined in Section~\ref{sec:results}.}
\label{tab:agg_domain}
\end{table}

\subsection{Performance Differences Across Tasks}
The clearest takeaway from these results is that the mean performance advantage of TNNs over GNNs is task-dependent. For example, it is modest on community detection and more pronounced on triangle counting.

On \textbf{community detection}, Track 2 attains a mean accuracy of $0.4343$ against $0.4241$ for Track 1, a modest descriptive difference of 1.02 percentage points. The best individual result comes from Track 1: GRIT (PR \#370) reaches $0.5207$, ahead of the best Track 2 model, HiGCN (PR \#364), at $0.4869$.

On \textbf{triangle counting}, the picture changes sharply. Track 2 attains a mean error of $0.3828$ against $0.7097$ for Track 1; the Track 1 mean is 1.85 times the Track 2 mean. The best submission overall is a combinatorial model, CTNN (PR \#418), at $0.0024$---more than two orders of magnitude below the Track 1 mean. This is consistent with the motivation behind encoding higher-order structure. We note that the track comparison does not isolate the contribution of that structure from other architectural and implementation choices.

Table~\ref{tab:agg_domain} gives a more detailed view. On triangle counting, simplicial submissions have the lowest mean error ($0.2586$), followed by cellular ($0.3144$), combinatorial ($0.3551$), hypergraph ($0.6505$) and graph submissions ($0.7097$).

\subsection{Structural Information Beyond Track Labels}
Going beyond track aggregations, we find that several of the strongest triangle-counting submissions are Track 1 models. For example, TopoPolynormer (PR \#353, $0.0451$), ESC-GNN (PR \#419, $0.0552$), GSN (PR \#407, $0.0665$), GRIT (PR \#370, $0.0693$) and HOD-GNN (PR \#402, $0.0802$) all outperform the Track 2 mean, and several outperform the majority of Track 2 submissions. A common property of these architectures is that while they operate on graph-based domains, they inject substructure or structural-encoding information, such as subgraph counts, random-walk encodings, or explicit higher-order descriptors.

These results show that a graph-domain model equipped with structural information can perform competitively on triangle counting. Higher-order domains offer another way to supply structural information, but performance also varies substantially among TNN submissions. Establishing the separate contributions of the lifting, structural encoding and message-passing scheme via controlled ablations is a promising avenue for future work.

\subsection{Structural Sensitivity}
The controlled grid allows us to ask how each family responds to the structural axes.
Figure~\ref{fig:score_distributions_homophily} reveals substantial overlap between GNNs and TNNs on community detection. At high homophily, their median accuracies are close ($0.6170$ and $0.6161$, respectively), despite the larger difference between their means. On triangle counting, TNNs achieve lower median error across all three homophily regimes, showing that their advantage is also reflected in the medians. Nevertheless, performance varies considerably within each family, and several GNN submissions achieve errors below the TNN median in every homophily regime.

\paragraph{Homophily.} On community detection, both tracks lose accuracy as homophily drops: from $0.5765$ to $0.3088$ for Track 1 and from $0.5970$ to $0.3061$ for Track 2. The Track 2 mean exceeds the Track 1 mean by 2.04 percentage points at high homophily and 1.31 at mid homophily, while Track 1 leads by 0.28 percentage points at low homophily. Thus, the observed Track 2 advantage does not extend to the low-homophily regime on this task. On triangle counting, the direction is reversed: mean error decreases as homophily falls, from $1.4304$ to $0.0887$ for Track 1 and from $0.7805$ to $0.0428$ for Track 2.

\paragraph{Average degree and degree distribution.} Both tracks perform better on community detection at high average degree than at low ($0.4491$ versus $0.3991$ for Track 1; $0.4588$ versus $0.4099$ for Track 2), consistent with denser neighborhoods carrying more signal. On triangle counting the pattern inverts, with normalized error rising at high degree and at high homophily in both tracks. Heavy-tailed degree distributions (low $\gamma$) also yield higher mean normalized error than homogeneous ones in both tracks. 

\subsection{Leaderboard Position and Award Outcome}
\label{subsec:leaderboard_vs_award}
Because both a leaderboard and a set of awards appear in this paper for the first time, we make their relationship explicit. As described in Section~\ref{subsec:evaluation}, awards were decided by Condorcet vote on implementation quality, not on benchmark score, and the two rankings differ considerably: the Track 1 winner, DSNN (PR \#412), ranks 62nd of 65 on community detection, and the Track 2 co-winner TopoU-Net (PR \#372) ranks 59th. The awards recognize implementation quality and reusability; Tables~\ref{tab:lb_cd} and~\ref{tab:lb_tc} describe performance on the two benchmark tasks. Award outcomes and leaderboard positions should therefore be interpreted separately.

\section{Discussion \& Concluding Remarks}
\label{sec:discussion}
\label{sec:conclusion}

This edition set out to answer two questions: how specific graph properties affect classical GNNs versus higher-order TNNs, and under what data regimes TNNs provide capabilities that standard GNNs do not. The results above provide a first, qualified answer to both.

On the first question, the community-detection gap narrows as homophily falls, with a small reversal in favour of GNNs at low homophily; a higher-order domain alone therefore does not establish robustness in that regime. On the second question, the submitted Track 2 cohort has lower mean error on triangle counting, while its mean accuracy advantage on community detection is small. The strong triangle-counting results of several Track 1 models suggest that both higher-order architectures and structurally informed graph models merit consideration for this task. Extending this observation to other motifs or tasks requires further evaluation.

We highlight the finding that substructure-aware GNNs close much of the gap on triangle counting as it points to a possible reframing of the comparison. Rather than a strict dichotomy between two model families, the results motivate studying the structural information supplied to models in both tracks. Both communities therefore face the same practical question: how to give networks access to structure beyond edges.

Several limitations bound these conclusions. The benchmark comprises two tasks on one family of synthetic graphs; performance on real-world data with correlated structural properties may differ. The results concern the submitted configurations, and their relative performance may change under different tuning budgets. Selecting variants by reported test accuracy can favour PRs with more variants and make community-detection performance estimates optimistic. Implementation quality, while reviewed, inevitably varies. Moreover, the aggregate statistics we report summarize a heterogeneous population of architectures, so track-level means should be interpreted as descriptions of the submissions rather than defined properties of the underlying model classes. Extending the benchmark suite with additional tasks and re-running it with per-model tuning are the natural next steps. Both are straightforward within \texttt{TopoBench}.

The challenge yielded 68 valid submissions from 34 teams, enriching \texttt{TopoBench} with implementations from both the GNN and TDL literatures. It also produced the first cross-track comparison in this challenge series under a shared benchmarking protocol, supported by a public leaderboard. We hope this infrastructure hosted in \texttt{TopoBench}  will support research within the communities, facilitate comparisons between models, and encourage the exchange of methods and ideas between them.

\acks{We thank Arlequin AI and New Theory for sponsoring this edition of the challenge, the Geometric Intelligence Lab (UC Santa Barbara) and the IMOS Lab (EPFL) for supporting the internship opportunities, and the TAG-DS organizers for hosting the challenge. We are grateful to all \texttt{TopoBench} maintainers and collaborators who served on the review panel, and above all to the participants, whose contributions constitute the substance of this paper. N. M., G. B., and M. P. acknowledge partial support from NSF grant 2602079 and the Noyce foundation.}

\newpage
\bibliography{references}

@article{barbarossa2020topological,
  title={Topological signal processing over simplicial complexes},
  author={Barbarossa, Sergio and Sardellitti, Stefania},
  journal={IEEE Transactions on Signal Processing},
  volume={68},
  pages={2992--3007},
  year={2020},
  publisher={IEEE}
}

@misc{hajij2022topologicalDL,
  title = {{Topological Deep Learning: Going Beyond Graph Data}},
  author = {Mustafa Hajij and Ghada Zamzmi and Theodore Papamarkou and Nina Miolane and Aldo Guzm{\'a}n-S{\'a}enz and Karthikeyan {Natesan Ramamurthy} and Tolga Birdal and Tamal K. Dey and Soham Mukherjee and Shreyas N. Samaga and Neal Livesay and Robin Walters and Paul Rosen and Michael T. Schaub},
  howpublished = {arXiv preprint arXiv:2206.00606},
  year = {2022},
  url = {https://arxiv.org/abs/2206.00606}
}

@article{papillon2023architectures,
  title={Architectures of Topological Deep Learning: A Survey of Message-Passing Topological Neural Networks},
  author={Papillon, Mathilde and Sanborn, Sophia and Hajij, Mustafa and Miolane, Nina},
  journal={arXiv preprint arXiv:2304.10031},
  year={2023}
}

@article{bick2023higher,
  title={What are higher-order networks?},
  author={Bick, Christian and Gross, Elizabeth and Harrington, Heather A and Schaub, Michael T},
  journal={SIAM Review},
  volume={65},
  number={3},
  pages={686--731},
  year={2023},
  publisher={SIAM}
}

@article{
telyatnikov2025topobench,
title={TopoBench: A Framework for Benchmarking Topological Deep Learning},
author={Lev Telyatnikov and Guillermo Bernardez and Marco Montagna and Mustafa Hajij and Martin Carrasco and Pavlo Vasylenko and Mathilde Papillon and Ghada Zamzmi and Michael T Schaub and Jonas Verhellen and Pavel Snopov and Bertran Miquel-Oliver and Manel Gil-Sorribes and Alexis Molina and Victor Guallar and Theodore Long and Julian Suk and Patryk Rygiel and Alexander V Nikitin and Giordan Escalona and Michael Banf and Dominik Filipiak and Liliya Imasheva and Max Schattauer and Alvaro L. Martinez and Halley Fritze and Marissa Masden and Valentina S{\'a}nchez and Manuel Lecha and Andrea Cavallo and Claudio Battiloro and Matthew Piekenbrock and Mauricio Tec and George Dasoulas and Nina Miolane and Simone Scardapane and Theodore Papamarkou},
journal={Journal of Data-centric Machine Learning Research},
year={2025},
url={https://openreview.net/forum?id=07sTzyEVtY},
note={}
}

@phdthesis{bodnar_2022, 
    title={Topological Deep Learning: Graphs, Complexes, Sheaves}, 
    url={https://www.repository.cam.ac.uk/handle/1810/350982}, 
    DOI={10.17863/CAM.97212}, 
    school={Apollo - University of Cambridge Repository}, 
    author={Bodnar, Cristian}, 
    year={2022} 
}

@inproceedings{kipf2017semi,
author = {Kipf, Thomas N. and Welling, Max},
title = {{Semi-Supervised Classification with Graph Convolutional Networks}},
booktitle = {ICLR},
year = {2017}
}

@inproceedings{bernardez2024icml,
  title={ICML Topological Deep Learning Challenge 2024: Beyond the Graph Domain},
  author={Bern{\'a}rdez, Guillermo and Telyatnikov, Lev and Montagna, Marco and Baccini, Federica and Papillon, Mathilde and Ferriol-Galm{\'e}s, Miquel and Hajij, Mustafa and Papamarkou, Theodore and Bucarelli, Maria Sofia and Zaghen, Olga and others},
  booktitle={Geometry-grounded Representation Learning and Generative Modeling Workshop (GRaM) at ICML 2024},
  pages={420--428},
  year={2024},
  organization={PMLR}
}

@InProceedings{pmlr-v221-papillon23a,
  title = 	 {ICML 2023 Topological Deep Learning Challenge: Design and Results},
  author =       {Papillon, Mathilde and Hajij, Mustafa and Myers, Audun and Jenne, Helen and Mathe, Johan and Papamarkou, Theodore and Guzm\'{a}n-S\'{a}enz, Aldo and Livesay, Neal and Dey, Tamal and Rabinowitz, Abraham and Brent, Aiden and Salatiello, Alessandro and Nikitin, Alexander and Zia, Ali and Battiloro, Claudio and Gavrilev, Dmitrii and Magai, German and Bazhenov, Gleb and Bernardez, Guillermo and Spinelli, Indro and Agerberg, Jens and Nadimpalli, Kalyan and Telyatninkov, Lev and Scofano, Luca and Testa, Lucia and Lecha, Manuel and Yang, Maosheng and Hassanin, Mohammed and Gardaa, Odin Hoff and Zaghen, Olga and Hausner, Paul and Snopoff, Paul and Ballester, Rub\'{e}n and Barikbin, Sadrodin and Escalera, Sergio and Fiorellino, Simone and Kvinge, Henry and Ramamurthy, Karthikeyan Natesan and Rosen, Paul and Walters, Robin and Samaga, Shreyas N. and Mukherjee, Soham and Sanborn, Sophia and Emerson, Tegan and Doster, Timothy and Birdal, Tolga and Khamis, Abdelwahed and Scardapane, Simone and Singh, Suraj and Malygina, Tatiana and Yue, Yixiao and Miolane, Nina},
  booktitle = 	 {Proceedings of 2nd Annual Workshop on Topology, Algebra, and Geometry in Machine Learning (TAG-ML)},
  pages = 	 {3--8},
  year = 	 {2023},
  editor = 	 {Doster, Timothy and Emerson, Tegan and Kvinge, Henry and Miolane, Nina and Papillon, Mathilde and Rieck, Bastian and Sanborn, Sophia},
  volume = 	 {221},
  series = 	 {Proceedings of Machine Learning Research},
  month = 	 {28 Jul},
  publisher =    {PMLR},
  url = 	 {https://proceedings.mlr.press/v221/papillon23a.html}
}

@inproceedings{van2025graphuniverse,
  title={GraphUniverse: Synthetic graph generation for evaluating inductive generalization},
  author={Van Langendonck, Louis Emiel T and Bern{\'a}rdez, Guillermo and Miolane, Nina and Barlet-Ros, Pere},
  booktitle={International Conference on Learning Representations},
  volume={2026},
  pages={138611--138642},
  year={2026}
}

@article{hajij2024topox,
  title={TopoX: a suite of Python packages for machine learning on topological domains},
  author={Hajij, Mustafa and Papillon, Mathilde and Frantzen, Florian and Agerberg, Jens and AlJabea, Ibrahem and Ballester, Rub{\'e}n and Battiloro, Claudio and Bern{\'a}rdez, Guillermo and Birdal, Tolga and Brent, Aiden and others},
  journal={Journal of Machine Learning Research},
  volume={25},
  number={374},
  pages={1--8},
  year={2024}
}

@article{scarselli2008graph,
  title={The graph neural network model},
  author={Scarselli, Franco and Gori, Marco and Tsoi, Ah Chung and Hagenbuchner, Markus and Monfardini, Gabriele},
  journal={IEEE transactions on neural networks},
  volume={20},
  number={1},
  pages={61--80},
  year={2008},
  publisher={IEEE}
}

@InProceedings{bernardez2026challenge,
  title =      {Topological Deep Learning Challenge 2025: Expanding the Data Landscape},
  author =       {Bern\'ardez, Guillermo and Telyatnikov, Lev and Papillon, Mathilde and Montagna, Marco and Theiler, Raffael and Cornelis, Louisa and Mathe, Johan and Ferriol, Miquel and Vasylenko, Pavlo and Van Looy, Jan-Willem and Testa, Lucia and Neri, Bruno and Genovese, Donatella and Weber, Melanie and Wei, Amaury and Devoto, Alessio and Weers, Alexander and Jankowski, Robert and Cino, Loris and Leko, David and Banf, Michael and M\"{u}ller, Jonas and Grapentin, Thomas and Paik, Taejin and Dutta, Abhijeet and Walter, Hugo and Fontanari, Thomas Vaitses and Ghasemi, Ali and Loi, Dario and S\'aez de Oc\'ariz Borde, Haitz and Aguilar-Arg\"uello, Gabriela and da Rosa, Giovanni B. and Saulus, Th\'eo and Dolores-Cuenca, Eric Rubiel and Di Nino, Leonardo and Leroy, Pierrick and Pandolfo, Mario Edoardo and Cavallo, Andrea and Qin, Yu and Snopov, Pavel and Akbari, Amirreza and Meza-Ch\'avez, Ixchel and Van Langendonck, Louis and Able, Jared and Meshcheryakova, Maria Yuffa and Tsay, Henry and Beni\'c, Luka and Filipiak, Dominik and Liu, Patrick and Liang, Huidong and Santos da Rosa Jr., Alexsandro and Cattai, Tiziana and Borges, Henrique M. and Grimaldi, Enrico and Lecha, Manuel and Battiloro, Claudio and Liu, Xuan-Chen and Deshpande, Raj and Johnson, Graham and Morgunov, Igor and Micheron, Hugo and Devaux, R\'emi and Jardin, Antoine and Emerson, Tegan and Fink, Olga and Miolane, Nina},
  booktitle =      {Proceedings of the 1st Conference on Topology, Algebra, and Geometry in Data Science(TAG-DS 2025)},
  pages =      {4--14},
  year =      {2026},
  editor =      {Bernardez Gil, Guillermo and Black, Mitchell and Cloninger, Alexander and Doster, Timothy and Emerson, Tegan and Garc\'ia-Rodondo, Ines and Holtz, Chester and Kotak, Mit and Kvinge, Henry and Mishne, Gal and Papillon, Mathilde and Pouplin, Alison and Rainey, Katie and Rieck, Bastian and Telyatnikov, Lev and Yeats, Eric and Wang, Qingsong and Wang, Yusu and Wayland, Jeremy},
  volume =      {321},
  series =      {Proceedings of Machine Learning Research},
  month =      {01--02 Dec},
  publisher =    {PMLR},
  url =      {https://proceedings.mlr.press/v321/bernardez26a.html}
}

@misc{grover2026sheafneuralnetworksuse,
      title={Do Sheaf Neural Networks Use Holonomy? A Measure--Intervene--Control Study},
      author={Ankit Grover and Rémi Bourgerie},
      year={2026},
      eprint={2607.19514},
      archivePrefix={arXiv},
      primaryClass={cs.LG},
      url={https://arxiv.org/abs/2607.19514},
}

@misc{tdl26_ballester2024attendingtransformer,
  title = {{Attending to Topological Spaces: The Cellular Transformer}},
  author = {Rub{\'e}n Ballester and Pablo Hern{\'a}ndez-Garc{\'{\i}}a and Mathilde Papillon and Claudio Battiloro and Nina Miolane and Tolga Birdal and Carles Casacuberta and Sergio Escalera and Mustafa Hajij},
  howpublished = {arXiv preprint arXiv:2405.14094},
  year = {2024},
  url = {https://arxiv.org/abs/2405.14094}
}

@inproceedings{tdl26_bamberger2025bundlegraphs,
  title = {{Bundle Neural Network for message diffusion on graphs}},
  author = {Jacob Bamberger and Federico Barbero and Xiaowen Dong and Michael M. Bronstein},
  booktitle = {International Conference on Learning Representations},
  year = {2025},
  url = {https://openreview.net/forum?id=scI9307PLG}
}

@inproceedings{tdl26_barbero2022sheaflaplacians,
  title = {{Sheaf Neural Networks with Connection Laplacians}},
  author = {Federico Barbero and Cristian Bodnar and Haitz {S{\'a}ez de Oc{\'a}riz Borde} and Michael Bronstein and Petar Veli{\v{c}}kovi{\'c} and Pietro Li{\`o}},
  booktitle = {Proceedings of Topological, Algebraic, and Geometric Learning Workshops 2022},
  volume = {196},
  pages = {28--36},
  series = {Proceedings of Machine Learning Research},
  publisher = {PMLR},
  year = {2022},
  url = {https://proceedings.mlr.press/v196/barbero22a.html}
}

@misc{tdl26_barbero2022sheafnetworks,
  title = {{Sheaf Attention Networks}},
  author = {Federico Barbero and Cristian Bodnar and Haitz {S{\'a}ez de Oc{\'a}riz Borde} and Pietro Li{\`o}},
  howpublished = {NeurIPS 2022 Workshop on Symmetry and Geometry in Neural Representations},
  year = {2022},
  url = {https://openreview.net/forum?id=LIDvgVjpkZr}
}

@article{tdl26_battiloro2024generalizednetworks,
  title = {{Generalized Simplicial Attention Neural Networks}},
  author = {Claudio Battiloro and Lucia Testa and Lorenzo Giusti and Stefania Sardellitti and Paolo {Di Lorenzo} and Sergio Barbarossa},
  journal = {IEEE Transactions on Signal and Information Processing over Networks},
  volume = {10},
  pages = {833--850},
  publisher = {IEEE},
  year = {2024},
  doi = {10.1109/TSIPN.2024.3485473},
  url = {https://ieeexplore.ieee.org/document/10735150/}
}

@inproceedings{tdl26_battiloro2025enetworks,
  title = {{E(n) Equivariant Topological Neural Networks}},
  author = {Claudio Battiloro and Ege Karaismailo{\u{g}}lu and Mauricio Tec and George Dasoulas and Michelle Audirac and Francesca Dominici},
  booktitle = {International Conference on Learning Representations},
  year = {2025},
  url = {https://openreview.net/forum?id=Ax3uliEBVR}
}

@article{tdl26_bo2021beyondnetworks,
  title = {{Beyond Low-frequency Information in Graph Convolutional Networks}},
  author = {Deyu Bo and Xiao Wang and Chuan Shi and Huawei Shen},
  journal = {Proceedings of the AAAI Conference on Artificial Intelligence},
  volume = {35},
  number = {5},
  pages = {3950--3957},
  year = {2021},
  doi = {10.1609/aaai.v35i5.16514},
  url = {https://ojs.aaai.org/index.php/AAAI/article/view/16514}
}

@inproceedings{tdl26_bodnar2021weisfeilernetworks,
  title = {{Weisfeiler and Lehman Go Topological: Message Passing Simplicial Networks}},
  author = {Cristian Bodnar and Fabrizio Frasca and Yuguang Wang and Nina Otter and Guido F. Mont{\'u}far and Pietro Li{\`o} and Michael Bronstein},
  booktitle = {Proceedings of the 38th International Conference on Machine Learning},
  volume = {139},
  pages = {1026--1037},
  series = {Proceedings of Machine Learning Research},
  publisher = {PMLR},
  year = {2021},
  url = {https://proceedings.mlr.press/v139/bodnar21a.html}
}

@inproceedings{tdl26_bodnar2021weisfeilernetworksa,
  title = {{Weisfeiler and Lehman Go Cellular: CW Networks}},
  author = {Cristian Bodnar and Fabrizio Frasca and Nina Otter and Yuguang Wang and Pietro Li{\`o} and Guido F. Mont{\'u}far and Michael Bronstein},
  booktitle = {Advances in Neural Information Processing Systems},
  volume = {34},
  pages = {2625--2640},
  year = {2021},
  url = {https://proceedings.neurips.cc/paper/2021/hash/157792e4abb490f99dbd738483e0d2d4-Abstract.html}
}

@inproceedings{tdl26_bodnar2022neuralgnns,
  title = {{Neural Sheaf Diffusion: A Topological Perspective on Heterophily and Oversmoothing in GNNs}},
  author = {Cristian Bodnar and Francesco {Di Giovanni} and Benjamin Chamberlain and Pietro Li{\`o} and Michael Bronstein},
  booktitle = {Advances in Neural Information Processing Systems},
  volume = {35},
  pages = {18527--18541},
  year = {2022},
  doi = {10.52202/068431-1346},
  url = {https://proceedings.neurips.cc/paper_files/paper/2022/hash/75c45fca2aa416ada062b26cc4fb7641-Abstract-Conference.html}
}

@misc{tdl26_bourgerie2026deepdiffusion,
  title = {{Deep Neural Sheaf Diffusion}},
  author = {R{\'e}mi Bourgerie and {\v{S}}ar{\={u}}nas Girdzijauskas and Viktoria Fodor},
  howpublished = {arXiv preprint arXiv:2605.19021},
  year = {2026},
  url = {https://arxiv.org/abs/2605.19021}
}

@article{tdl26_bouritsas2023improvingcounting,
  title = {{Improving Graph Neural Network Expressivity via Subgraph Isomorphism Counting}},
  author = {Giorgos Bouritsas and Fabrizio Frasca and Stefanos Zafeiriou and Michael M. Bronstein},
  journal = {IEEE Transactions on Pattern Analysis and Machine Intelligence},
  volume = {45},
  number = {1},
  pages = {657--668},
  publisher = {IEEE},
  year = {2023},
  doi = {10.1109/TPAMI.2022.3154319},
  url = {https://ieeexplore.ieee.org/document/9721082/}
}

@misc{tdl26_caralt2026necessitylaplacians,
  title = {{On the Necessity of Learnable Sheaf Laplacians}},
  author = {Ferran {Hernandez Caralt} and Mar {Gonz{\`a}lez i Catal{\`a}} and Adri{\'a}n Bazaga and Pietro Li{\`o}},
  howpublished = {arXiv preprint arXiv:2603.05395},
  year = {2026},
  url = {https://arxiv.org/abs/2603.05395}
}

@inproceedings{tdl26_chien2021adaptivenetwork,
  title = {{Adaptive Universal Generalized PageRank Graph Neural Network}},
  author = {Eli Chien and Jianhao Peng and Pan Li and Olgica Milenkovic},
  booktitle = {International Conference on Learning Representations},
  year = {2021},
  url = {https://openreview.net/forum?id=n6jl7fLxrP}
}

@inproceedings{tdl26_choi2023greadnetworks,
  title = {{GREAD: Graph Neural Reaction-Diffusion Networks}},
  author = {Jeongwhan Choi and Seoyoung Hong and Noseong Park and Sung-Bae Cho},
  booktitle = {Proceedings of the 40th International Conference on Machine Learning},
  volume = {202},
  pages = {5722--5747},
  series = {Proceedings of Machine Learning Research},
  publisher = {PMLR},
  year = {2023},
  url = {https://proceedings.mlr.press/v202/choi23a.html}
}

@inproceedings{tdl26_deng2024polynormertime,
  title = {{Polynormer: Polynomial-Expressive Graph Transformer in Linear Time}},
  author = {Chenhui Deng and Zichao Yue and Zhiru Zhang},
  booktitle = {International Conference on Learning Representations},
  year = {2024},
  url = {https://openreview.net/forum?id=hmv1LpNfXa}
}

@inproceedings{tdl26_duta2023sheafnetworks,
  title = {{Sheaf Hypergraph Networks}},
  author = {Iulia Duta and Giulia Cassar{\`a} and Fabrizio Silvestri and Pietro Li{\`o}},
  booktitle = {Advances in Neural Information Processing Systems},
  volume = {36},
  pages = {12087--12099},
  year = {2023},
  doi = {10.52202/075280-0529},
  url = {https://proceedings.neurips.cc/paper_files/paper/2023/hash/27f243af2887d7f248f518d9b967a882-Abstract-Conference.html}
}

@misc{tdl26_duta2025wassersteinnetwork,
  title = {{Wasserstein Hypergraph Neural Network}},
  author = {Iulia Duta and Pietro Li{\`o}},
  howpublished = {arXiv preprint arXiv:2506.09682},
  year = {2025},
  url = {https://arxiv.org/abs/2506.09682}
}

@misc{tdl26_ebli2020simplicialnetworks,
  title = {{Simplicial Neural Networks}},
  author = {Stefania Ebli and Micha{\"e}l Defferrard and Gard Spreemann},
  howpublished = {arXiv preprint arXiv:2010.03633},
  year = {2020},
  url = {https://arxiv.org/abs/2010.03633}
}

@inproceedings{tdl26_einizade2025continuousnetworks,
  title = {{Continuous Simplicial Neural Networks}},
  author = {Aref Einizade and Dorina Thanou and Fragkiskos D. Malliaros and Jhony H. Giraldo},
  booktitle = {Advances in Neural Information Processing Systems},
  volume = {38},
  pages = {59467--59495},
  year = {2025},
  url = {https://proceedings.neurips.cc/paper_files/paper/2025/hash/55c944c9d908af245fa658b0e02b9003-Abstract-Conference.html}
}

@inproceedings{tdl26_eitan2025topologicalexpressivity,
  title = {{Topological Blindspots: Understanding and Extending Topological Deep Learning Through the Lens of Expressivity}},
  author = {Yam Eitan and Yoav Gelberg and Guy Bar-Shalom and Fabrizio Frasca and Michael M. Bronstein and Haggai Maron},
  booktitle = {International Conference on Learning Representations},
  year = {2025},
  url = {https://openreview.net/forum?id=EzjsoomYEb}
}

@inproceedings{tdl26_eitan2026expressivederivatives,
  title = {{On The Expressive Power of GNN Derivatives}},
  author = {Yam Eitan and Moshe Eliasof and Yoav Gelberg and Fabrizio Frasca and Guy Bar-Shalom and Haggai Maron},
  booktitle = {International Conference on Learning Representations},
  year = {2026},
  url = {https://openreview.net/forum?id=Tvat33IDmK}
}

@inproceedings{tdl26_ekbote2023figureaugmentations,
  title = {{FiGURe: Simple and Efficient Unsupervised Node Representations with Filter Augmentations}},
  author = {Chanakya Ekbote and Ajinkya Deshpande and Arun Iyer and Sundararajan Sellamanickam and Ramakrishna Bairi},
  booktitle = {Advances in Neural Information Processing Systems},
  volume = {36},
  year = {2023},
  url = {https://proceedings.neurips.cc/paper_files/paper/2023/hash/6f479ea488e0908ac8b1b37b27fd134c-Abstract-Conference.html}
}

@inproceedings{tdl26_finkelshtein2024cooperativenetworks,
  title = {{Cooperative Graph Neural Networks}},
  author = {Ben Finkelshtein and Xingyue Huang and Michael M. Bronstein and Ismail Ilkan Ceylan},
  booktitle = {Proceedings of the 41st International Conference on Machine Learning},
  volume = {235},
  pages = {13633--13659},
  series = {Proceedings of Machine Learning Research},
  publisher = {PMLR},
  year = {2024},
  url = {https://proceedings.mlr.press/v235/finkelshtein24a.html}
}

@inproceedings{tdl26_fiorellino2024topologicalair,
  title = {{Topological Neural Networks over the Air}},
  author = {Simone Fiorellino and Claudio Battiloro and Paolo {Di Lorenzo}},
  booktitle = {ICASSP 2024 - 2024 IEEE International Conference on Acoustics, Speech and Signal Processing (ICASSP)},
  pages = {12986--12990},
  publisher = {IEEE},
  year = {2024},
  doi = {10.1109/ICASSP48485.2024.10446693},
  url = {https://ieeexplore.ieee.org/document/10446693}
}

@inproceedings{tdl26_fiorini2026sheavesawakening,
  title = {{Sheaves Reloaded: A Direction Awakening}},
  author = {Stefano Fiorini and Hakan Aktas and Iulia Duta and Pietro Morerio and Alessio {Del Bue} and Pietro Li{\`o} and Stefano Coniglio},
  booktitle = {International Conference on Learning Representations},
  year = {2026},
  url = {https://openreview.net/forum?id=iDiiETH7Qv}
}

@inproceedings{tdl26_franco2026differentiablenetworks,
  title = {{Differentiable Lifting for Topological Neural Networks}},
  author = {Jorge Luiz Franco and Gabriel Duarte and Alexander V Nikitin and Moacir A Ponti and Diego Mesquita and Amauri H Souza},
  booktitle = {International Conference on Learning Representations},
  year = {2026},
  url = {https://openreview.net/forum?id=eC89CbINIw}
}

@misc{tdl26_gaurav2026topoudomains,
  title = {{TopoU-Net: A U-Net Architecture for Topological Domains}},
  author = {Gaurav Gaurav and Ibrahem ALJabea and Yaroslav Zakomornyy and Eric Frank and Mohamed Elhamdadi and Theodore Papamarkou and Mustafa Hajij},
  howpublished = {arXiv preprint arXiv:2605.10091},
  year = {2026},
  url = {https://arxiv.org/abs/2605.10091}
}

@inproceedings{tdl26_ghazizadeh2025gammagnns,
  title = {{GAMMA: Gated Multi-hop Message Passing for Homophily-Agnostic Node Representation in GNNs}},
  author = {Amir Ghazizadeh and Rickard Ewetz and Hao Zheng},
  booktitle = {Advances in Neural Information Processing Systems},
  volume = {38},
  pages = {11726--11777},
  year = {2025},
  doi = {10.52202/085713-0393},
  url = {https://proceedings.neurips.cc/paper_files/paper/2025/hash/1129f729097a28bfb5b836ec4bf94478-Abstract-Conference.html}
}

@inproceedings{tdl26_guo2023graphbases,
  title = {{Graph Neural Networks with Learnable and Optimal Polynomial Bases}},
  author = {Yuhe Guo and Zhewei Wei},
  booktitle = {Proceedings of the 40th International Conference on Machine Learning},
  volume = {202},
  pages = {12077--12097},
  series = {Proceedings of Machine Learning Research},
  publisher = {PMLR},
  year = {2023},
  url = {https://proceedings.mlr.press/v202/guo23i.html}
}

@inproceedings{tdl26_hajij2025copresheafframework,
  title = {{Copresheaf Topological Neural Networks: A Generalized Deep Learning Framework}},
  author = {Mustafa Hajij and Lennart Bastian and Sarah Osentoski and Hardik Kabaria and John L. Davenport and Sheik Dawood and Balaji Cherukuri and Joseph G. Kocheemoolayil and Nastaran Shahmansouri and Adrian Lew and Theodore Papamarkou and Tolga Birdal},
  booktitle = {Advances in Neural Information Processing Systems},
  volume = {38},
  pages = {149759--149803},
  year = {2025},
  doi = {10.52202/085713-5009},
  url = {https://proceedings.neurips.cc/paper_files/paper/2025/hash/dc62cd4ec77f3bbec8e6245d0bd91d08-Abstract-Conference.html}
}

@article{tdl26_huang2024highercomplexes,
  title = {{Higher-Order Graph Convolutional Network with Flower-Petals Laplacians on Simplicial Complexes}},
  author = {Yiming Huang and Yujie Zeng and Qiang Wu and Linyuan L{\"u}},
  journal = {Proceedings of the AAAI Conference on Artificial Intelligence},
  volume = {38},
  number = {11},
  pages = {12653--12661},
  publisher = {Association for the Advancement of Artificial Intelligence},
  year = {2024},
  doi = {10.1609/aaai.v38i11.29160},
  url = {https://ojs.aaai.org/index.php/AAAI/article/view/29160}
}

@inproceedings{tdl26_kim2022equivariantnetworks,
  title = {{Equivariant Hypergraph Neural Networks}},
  author = {Jinwoo Kim and Saeyoon Oh and Sungjun Cho and Seunghoon Hong},
  booktitle = {Computer Vision -- ECCV 2022: 17th European Conference, Tel Aviv, Israel, October 23--27, 2022, Proceedings, Part XXI},
  volume = {13681},
  pages = {86--103},
  series = {Lecture Notes in Computer Science},
  publisher = {Springer, Cham},
  year = {2022},
  doi = {10.1007/978-3-031-19803-8_6},
  url = {https://link.springer.com/chapter/10.1007/978-3-031-19803-8_6}
}

@inproceedings{tdl26_lecha2025highernetworks,
  title = {{Higher-Order Topological Directionality and Directed Simplicial Neural Networks}},
  author = {Manuel Lecha and Andrea Cavallo and Francesca Dominici and Elvin Isufi and Claudio Battiloro},
  booktitle = {ICASSP 2025 - 2025 IEEE International Conference on Acoustics, Speech and Signal Processing (ICASSP)},
  pages = {1--5},
  publisher = {IEEE},
  year = {2025},
  doi = {10.1109/ICASSP49660.2025.10890264},
  url = {https://doi.org/10.1109/ICASSP49660.2025.10890264}
}

@article{tdl26_li2024generalizedgraphs,
  title = {{A Generalized Neural Diffusion Framework on Graphs}},
  author = {Yibo Li and Xiao Wang and Hongrui Liu and Chuan Shi},
  journal = {Proceedings of the AAAI Conference on Artificial Intelligence},
  volume = {38},
  number = {8},
  pages = {8707--8715},
  year = {2024},
  doi = {10.1609/aaai.v38i8.28716},
  url = {https://ojs.aaai.org/index.php/AAAI/article/view/28716}
}

@inproceedings{tdl26_lim2021largemethods,
  title = {{Large Scale Learning on Non-Homophilous Graphs: New Benchmarks and Strong Simple Methods}},
  author = {Derek Lim and Felix Hohne and Xiuyu Li and Sijia Linda Huang and Vaishnavi Gupta and Omkar Bhalerao and Ser Nam Lim},
  booktitle = {Advances in Neural Information Processing Systems},
  volume = {34},
  year = {2021},
  url = {https://proceedings.neurips.cc/paper/2021/hash/ae816a80e4c1c56caa2eb4e1819cbb2f-Abstract.html}
}

@inproceedings{tdl26_ma2023graphpassing,
  title = {{Graph Inductive Biases in Transformers without Message Passing}},
  author = {Liheng Ma and Chen Lin and Derek Lim and Adriana Romero-Soriano and Puneet K. Dokania and Mark Coates and Philip Torr and Ser-Nam Lim},
  booktitle = {Proceedings of the 40th International Conference on Machine Learning},
  volume = {202},
  pages = {23321--23337},
  series = {Proceedings of Machine Learning Research},
  publisher = {PMLR},
  year = {2023},
  url = {https://proceedings.mlr.press/v202/ma23c.html}
}

@inproceedings{tdl26_mule2026directionalhypergraphs,
  title = {{Directional Sheaf Hypergraph Networks: Unifying Learning on Directed and Undirected Hypergraphs}},
  author = {Emanuele Mule and Stefano Fiorini and Antonio Purificato and Federico Siciliano and Stefano Coniglio and Fabrizio Silvestri},
  booktitle = {International Conference on Learning Representations},
  year = {2026},
  url = {https://openreview.net/forum?id=d1hQkcDy7c}
}

@inproceedings{tdl26_mustafa2024gateneighbors,
  title = {{GATE: How to Keep Out Intrusive Neighbors}},
  author = {Nimrah Mustafa and Rebekka Burkholz},
  booktitle = {Proceedings of the 41st International Conference on Machine Learning},
  volume = {235},
  pages = {36990--37015},
  series = {Proceedings of Machine Learning Research},
  publisher = {PMLR},
  year = {2024},
  url = {https://proceedings.mlr.press/v235/mustafa24a.html}
}

@inproceedings{tdl26_paolino2024weisfeilerlearning,
  title = {{Weisfeiler and Leman Go Loopy: A New Hierarchy for Graph Representational Learning}},
  author = {Raffaele Paolino and Sohir Maskey and Pascal Welke and Gitta Kutyniok},
  booktitle = {Advances in Neural Information Processing Systems},
  volume = {37},
  pages = {120780--120831},
  year = {2024},
  doi = {10.52202/079017-3838},
  url = {https://proceedings.neurips.cc/paper_files/paper/2024/hash/dad28e90cd2c8caedf362d49c4d99e70-Abstract-Conference.html}
}

@inproceedings{tdl26_pei2024multimixing,
  title = {{Multi-Track Message Passing: Tackling Oversmoothing and Oversquashing in Graph Learning via Preventing Heterophily Mixing}},
  author = {Hongbin Pei and Yu Li and Huiqi Deng and Jingxin Hai and Pinghui Wang and Jie Ma and Jing Tao and Yuheng Xiong and Xiaohong Guan},
  booktitle = {Proceedings of the 41st International Conference on Machine Learning},
  volume = {235},
  pages = {40078--40091},
  series = {Proceedings of Machine Learning Research},
  publisher = {PMLR},
  year = {2024},
  url = {https://proceedings.mlr.press/v235/pei24a.html}
}

@inproceedings{tdl26_saxena2024dphgnnnetworks,
  title = {{DPHGNN: A Dual Perspective Hypergraph Neural Networks}},
  author = {Siddhant Saxena and Shounak Ghatak and Raghu Kolla and Debashis Mukherjee and Tanmoy Chakraborty},
  booktitle = {Proceedings of the 30th ACM SIGKDD Conference on Knowledge Discovery and Data Mining},
  pages = {2548--2559},
  publisher = {ACM},
  year = {2024},
  doi = {10.1145/3637528.3672047},
  url = {https://dl.acm.org/doi/10.1145/3637528.3672047}
}

@inproceedings{tdl26_southern2025balancingcentrality,
  title = {{Balancing Efficiency and Expressiveness: Subgraph GNNs with Walk-Based Centrality}},
  author = {Joshua Southern and Yam Eitan and Guy Bar-Shalom and Michael M. Bronstein and Haggai Maron and Fabrizio Frasca},
  booktitle = {Proceedings of the 42nd International Conference on Machine Learning},
  volume = {267},
  pages = {56615--56649},
  series = {Proceedings of Machine Learning Research},
  publisher = {PMLR},
  year = {2025},
  url = {https://proceedings.mlr.press/v267/southern25a.html}
}

@misc{tdl26_suk2022surfingsheaf,
  title = {{Surfing on the Neural Sheaf}},
  author = {Julian Suk and Lorenzo Giusti and Tamir Hemo and Miguel Lopez and Konstantinos Barmpas and Cristian Bodnar},
  howpublished = {NeurIPS 2022 Workshop on Symmetry and Geometry in Neural Representations},
  year = {2022},
  url = {https://openreview.net/forum?id=xOXFkyRzTlu}
}

@inproceedings{tdl26_sun2026arebundles,
  title = {{Are Common Substructures Transferable? Riemannian Graph Foundation Model with Neural Vector Bundles}},
  author = {Li Sun and Zhenhao Huang and Yiding Wang and Qin Chen and Pietro Li{\`o} and Philip S. Yu},
  booktitle = {Proceedings of the 43rd International Conference on Machine Learning},
  volume = {306},
  pages = {116668--116694},
  series = {Proceedings of Machine Learning Research},
  publisher = {PMLR},
  year = {2026},
  url = {https://proceedings.mlr.press/v306/sun26o.html}
}

@misc{tdl26_tandon2026consistentsheaves,
  title = {{Consistent Geometric Deep Learning via Hilbert Bundles and Cellular Sheaves}},
  author = {Kartik Tandon and Julian Gould and Tanishq Bhatia and Francesca Dominici and Alejandro Ribeiro and Claudio Battiloro},
  howpublished = {arXiv preprint arXiv:2605.06395},
  year = {2026},
  url = {https://arxiv.org/abs/2605.06395}
}

@inproceedings{tdl26_verma2024topologicalcontinuous,
  title = {{Topological Neural Networks go Persistent, Equivariant, and Continuous}},
  author = {Yogesh Verma and Amauri H Souza and Vikas Garg},
  booktitle = {Proceedings of the 41st International Conference on Machine Learning},
  volume = {235},
  pages = {49388--49407},
  series = {Proceedings of Machine Learning Research},
  publisher = {PMLR},
  year = {2024},
  url = {https://proceedings.mlr.press/v235/verma24a.html}
}

@inproceedings{tdl26_wang2022hownetworks,
  title = {{How Powerful are Spectral Graph Neural Networks}},
  author = {Xiyuan Wang and Muhan Zhang},
  booktitle = {Proceedings of the 39th International Conference on Machine Learning},
  volume = {162},
  pages = {23341--23362},
  series = {Proceedings of Machine Learning Research},
  publisher = {PMLR},
  year = {2022},
  url = {https://proceedings.mlr.press/v162/wang22am.html}
}

@inproceedings{tdl26_wang2023fromnetworks,
  title = {{From Hypergraph Energy Functions to Hypergraph Neural Networks}},
  author = {Yuxin Wang and Quan Gan and Xipeng Qiu and Xuanjing Huang and David Wipf},
  booktitle = {Proceedings of the 40th International Conference on Machine Learning},
  volume = {202},
  pages = {35605--35623},
  series = {Proceedings of Machine Learning Research},
  publisher = {PMLR},
  year = {2023},
  url = {https://proceedings.mlr.press/v202/wang23d.html}
}

@inproceedings{tdl26_wang2024gftvocabulary,
  title = {{GFT: Graph Foundation Model with Transferable Tree Vocabulary}},
  author = {Zehong Wang and Zheyuan Zhang and Nitesh V Chawla and Chuxu Zhang and Yanfang Ye},
  booktitle = {Advances in Neural Information Processing Systems},
  volume = {37},
  year = {2024},
  doi = {10.52202/079017-3412},
  url = {https://proceedings.neurips.cc/paper_files/paper/2024/hash/c23ccf9eedf87e4380e92b75b24955bb-Abstract-Conference.html}
}

@inproceedings{tdl26_wu2023sgformerrepresentations,
  title = {{SGFormer: Simplifying and Empowering Transformers for Large-Graph Representations}},
  author = {Qitian Wu and Wentao Zhao and Chenxiao Yang and Hengrui Zhang and Fan Nie and Haitian Jiang and Yatao Bian and Junchi Yan},
  booktitle = {Advances in Neural Information Processing Systems},
  volume = {36},
  year = {2023},
  doi = {10.52202/075280-2826},
  url = {https://proceedings.neurips.cc/paper_files/paper/2023/hash/cc57fac10eacadb3b72a907ac48f9a98-Abstract.html}
}

@inproceedings{tdl26_wu2025superchargingdiffusion,
  title = {{Supercharging Graph Transformers with Advective Diffusion}},
  author = {Qitian Wu and Chenxiao Yang and Kaipeng Zeng and Michael M. Bronstein},
  booktitle = {Proceedings of the 42nd International Conference on Machine Learning},
  volume = {267},
  pages = {67959--67985},
  series = {Proceedings of Machine Learning Research},
  publisher = {PMLR},
  year = {2025},
  url = {https://proceedings.mlr.press/v267/wu25an.html}
}

@inproceedings{tdl26_yadati2019hypergcnhypergraphs,
  title = {{HyperGCN: A New Method For Training Graph Convolutional Networks on Hypergraphs}},
  author = {Naganand Yadati and Madhav Nimishakavi and Prateek Yadav and Vikram Nitin and Anand Louis and Partha Talukdar},
  booktitle = {Advances in Neural Information Processing Systems},
  volume = {32},
  year = {2019},
  url = {https://proceedings.neurips.cc/paper/2019/hash/1efa39bcaec6f3900149160693694536-Abstract.html}
}

@inproceedings{tdl26_yan2024efficientpower,
  title = {{An Efficient Subgraph GNN with Provable Substructure Counting Power}},
  author = {Zuoyu Yan and Junru Zhou and Liangcai Gao and Zhi Tang and Muhan Zhang},
  booktitle = {Proceedings of the 30th ACM SIGKDD Conference on Knowledge Discovery and Data Mining},
  pages = {3702--3713},
  publisher = {ACM},
  year = {2024},
  doi = {10.1145/3637528.3671731},
  url = {https://dl.acm.org/doi/10.1145/3637528.3671731}
}

@article{tdl26_yan2025binarizednetworks,
  title = {{Binarized Simplicial Convolutional Neural Networks}},
  author = {Yi Yan and Ercan Engin Kuruoglu},
  journal = {Neural Networks},
  volume = {183},
  pages = {106928},
  publisher = {Elsevier},
  year = {2025},
  doi = {10.1016/j.neunet.2024.106928},
  url = {https://www.sciencedirect.com/science/article/abs/pii/S0893608024008578}
}

@article{tdl26_zhang2025graphclassification,
  title = {{A graph transformer with optimized attention scores for node classification}},
  author = {Yu Zhang and Xin Li and Yaoqun Xu and Xitong Xu and Zhe Wang},
  journal = {Scientific Reports},
  volume = {15},
  pages = {30015},
  year = {2025},
  doi = {10.1038/s41598-025-15551-2},
  url = {https://www.nature.com/articles/s41598-025-15551-2}
}

@inproceedings{tdl26_zhang2025herofilterrelations,
  title = {{HeroFilter: Adaptive Spectral Graph Filter for Varying Heterophilic Relations}},
  author = {Shuaicheng Zhang and Haohui Wang and Junhong Lin and Xiaojie Guo and Yada Zhu and Si Zhang and Dongqi Fu and Dawei Zhou},
  booktitle = {Advances in Neural Information Processing Systems},
  volume = {38},
  year = {2025},
  url = {https://openreview.net/forum?id=77Kw7b7EYb}
}

\newpage
\appendix

\section{Teams and Co-authors}
\label{app:teams}

Table~\ref{tab:teams} associates each participating team with the individual co-authors of this paper who belong to it, and lists the pull requests and awards of each team. Participants who submitted under more than one team appear once per team.

\begin{table}[htbp]
\centering
\scriptsize
\setlength{\tabcolsep}{4pt}
\begin{tabular}{@{}>{\raggedright\arraybackslash}p{0.18\textwidth} >{\raggedright\arraybackslash}p{0.28\textwidth} l >{\raggedright\arraybackslash}p{0.16\textwidth} >{\raggedright\arraybackslash}p{0.25\textwidth}@{}}
\toprule
\textbf{Team} & \textbf{Co-authors} & \textbf{Track} & \textbf{PR numbers} & \textbf{Award} \\
\midrule
AdamGohain & Adam Gohain & T1 & 359 &  \\
CoPreSheafers & Christine Zeh, Lennart Bastian & T2 & 418 &  \\
CtrlAltMe & Salwa Khattami & T2 & 397 &  \\
DeveshKaushal & Devesh Kaushal & T2 & 392 &  \\
E(n)igma & Gaurav Khanal & T2 & 320, 362, 391 & T2 joint 1st (ETNN-CoordinatePolicy) \\
Fernando Espinosa & Fernando David Espinosa I\~niguez & T2 & 410 &  \\
Francesco Orsi & Francesco Orsi & T2 & 393, 394 &  \\
GiggioIlGriggio & Alessio Comparini & T2 & 376 &  \\
Grold & Grold Otieno Mboya & T2 & 350 &  \\
howyadoin & Dario Loi, Gabriele Onorato & T1+T2 & 322, 326, 405, 408, 412 & T1 1st (DSNN); HM portfolio \\
Kajolya & Kolya Lettl, Kajal Singla & T1 & 407 &  \\
kxdenz & Kaden Zheng & T1 & 419 &  \\
LangDiff & Jonas M\"uller & T1 & 335, 399, 400, 420 &  \\
Lettlini & Kolya Lettl, Ghazal Mostofi & T1 & 396 &  \\
LoopCounter & Sud Athreya & T2 & 364, 365, 366, 367 &  \\
MISN\_AY & Mohammad Yusuf, Ankit Kumar & T1+T2 & 382, 383, 387, 388 &  \\
One Ring to Lift Them All & Jorge Luiz Franco & T2 & 406 & HM differentiable liftings \\
Oversmooth operators & Gabriel Franco, Yelizaveta Falkouskaya & T1+T2 & 398, 404 &  \\
Perelyn Research & Michael Banf, Dominik Filipiak & T1 & 361 &  \\
Prasanna28\-Devadiga & Prasanna Devadiga & T1 & 327 &  \\
PushparajD & Pushparaj Devadiga & T1 & 352, 355 &  \\
r2 & Anier Velasco Sotomayor & T1+T2 & 329, 354, 360, 414 &  \\
Raj Patil & Raj Patil & T1 & 411 &  \\
rayquaza & Luigi Petti, Eric Dolores-Cuenca & T2 & 413 &  \\
Remy & Stefano Blando & T2 & 417 &  \\
s/pairwise/ho & Valentina S\'anchez & T2 & 321 &  \\
Sheafu & Ankit Grover, R\'emi Bourgerie & T1 & 373, 374, 375 & HM supplementary analysis \\
SweetLesson & Aarav Gupta, Riyan Patel & T1 & 345, 353, 402 & T1 joint 2nd (HOD-GNN) \\
TJPaik & Taejin Paik & T1+T2 & 328, 332, 333, 334 &  \\
TopoCoop & Marcos Ch\'avez Mesa, Gabriel Chaviano D\'iaz & T1+T2 & 369, 371 &  \\
topolab & Chenguang Xu & T1+T2 & 415, 416 &  \\
TripleA & Amirreza Akbari, Alireza Mohammadian & T1+T2 & 363, 370, 372, 378, 384 & T1 joint 2nd (GRIT); T2 joint 1st (TopoU-Net); HM portfolio \\
weeyev & Vihan Tiwari & T1 & 421 &  \\
yeahkitory & Hisataka Kajita & T1 & 319 &  \\
\bottomrule
\end{tabular}
\caption{Teams and their co-authors, in alphabetical order of team name. T1 and T2 denote Track~1 (GNN) and Track~2 (TNN); HM denotes an honorable mention (Section~\ref{sec:submissions}). A dash marks a team for which no author information was provided.}
\label{tab:teams}
\end{table}

\section{Full Leaderboards}
\label{app:leaderboards}

Tables~\ref{tab:lb_cd} and~\ref{tab:lb_tc} give the rankings of the 65 entries included in the comparative analysis, with variant selection and exclusions as described in Section~\ref{sec:results}. Scores are means over the 12 GraphUniverse settings and three training seeds per setting; rankings use unrounded scores.

{\footnotesize
\setlength{\LTcapwidth}{\textwidth}
\begin{longtable}{@{}r >{\raggedright\arraybackslash}p{0.40\textwidth} l >{\raggedright\arraybackslash}p{0.20\textwidth} r@{}}
\toprule
\textbf{\#} & \textbf{Model} & \textbf{Track} & \textbf{Team} & \textbf{Score $\uparrow$} \\
\midrule
\endfirsthead
\toprule
\textbf{\#} & \textbf{Model} & \textbf{Track} & \textbf{Team} & \textbf{Score $\uparrow$} \\
\midrule
\endhead
\bottomrule
\caption{Community-detection leaderboard for the 65-entry analysis cohort (accuracy, higher is better; Section~\ref{sec:results}). Ranks use full-precision means.}\label{tab:lb_cd}\\
\endlastfoot
1 & \href{https://github.com/geometric-intelligence/TopoBench/pull/370}{Graph Inductive Bias Transformer (GRIT)} & T1 $\cdot$ GNN & TripleA & 0.5207 \\
2 & \href{https://github.com/geometric-intelligence/TopoBench/pull/360}{Filter Bank GNN (FiGURe)} & T1 $\cdot$ GNN & r2 & 0.4986 \\
3 & \href{https://github.com/geometric-intelligence/TopoBench/pull/354}{PolyFilter GNN (Favard)} & T1 $\cdot$ GNN & r2 & 0.4977 \\
4 & \href{https://github.com/geometric-intelligence/TopoBench/pull/364}{HiGCN} & T2 $\cdot$ TNN & LoopCounter & 0.4869 \\
5 & \href{https://github.com/geometric-intelligence/TopoBench/pull/415}{GREAD} & T1 $\cdot$ GNN & topolab & 0.4816 \\
6 & \href{https://github.com/geometric-intelligence/TopoBench/pull/353}{TopoPolynormer} & T1 $\cdot$ GNN & SweetLesson & 0.4798 \\
7 & \href{https://github.com/geometric-intelligence/TopoBench/pull/375}{Deep Neural Sheaf Diffusion (DNSD)} & T1 $\cdot$ GNN & Sheafu & 0.4743 \\
8 & \href{https://github.com/geometric-intelligence/TopoBench/pull/419}{ESC-GNN} & T1 $\cdot$ GNN & kxdenz & 0.4743 \\
9 & \href{https://github.com/geometric-intelligence/TopoBench/pull/376}{MPSN} & T2 $\cdot$ TNN & GiggioIlGriggio & 0.4742 \\
10 & \href{https://github.com/geometric-intelligence/TopoBench/pull/410}{SheafTSP} & T2 $\cdot$ TNN & Fernando Espinosa & 0.4735 \\
11 & \href{https://github.com/geometric-intelligence/TopoBench/pull/321}{SheafHyperGNN} & T2 $\cdot$ TNN & s/pairwise/ho & 0.4728 \\
12 & \href{https://github.com/geometric-intelligence/TopoBench/pull/406}{DiffLift-SMCN} & T2 $\cdot$ TNN & One Ring to Lift Them All & 0.4719 \\
13 & \href{https://github.com/geometric-intelligence/TopoBench/pull/352}{Generalized PageRank GNN} & T1 $\cdot$ GNN & PushparajD & 0.4697 \\
14 & \href{https://github.com/geometric-intelligence/TopoBench/pull/400}{MTGCN} & T1 $\cdot$ GNN & LangDiff & 0.4697 \\
15 & \href{https://github.com/geometric-intelligence/TopoBench/pull/418}{CTNN (copresheaf, gated)} & T2 $\cdot$ TNN & CoPreSheafers & 0.4690 \\
16 & \href{https://github.com/geometric-intelligence/TopoBench/pull/335}{HiD-Net} & T1 $\cdot$ GNN & LangDiff & 0.4685 \\
17 & \href{https://github.com/geometric-intelligence/TopoBench/pull/363}{Cell Isomorphism Network (CIN)} & T2 $\cdot$ TNN & TripleA & 0.4681 \\
18 & \href{https://github.com/geometric-intelligence/TopoBench/pull/383}{JacobiConv} & T1 $\cdot$ GNN & MISN\_AY & 0.4649 \\
19 & \href{https://github.com/geometric-intelligence/TopoBench/pull/404}{$r$-$\ell$MPNN} & T1 $\cdot$ GNN & Oversmooth operators & 0.4648 \\
20 & \href{https://github.com/geometric-intelligence/TopoBench/pull/398}{DPHGNN} & T2 $\cdot$ TNN & Oversmooth operators & 0.4641 \\
21 & \href{https://github.com/geometric-intelligence/TopoBench/pull/408}{DSHN} & T2 $\cdot$ TNN & howyadoin & 0.4633 \\
22 & \href{https://github.com/geometric-intelligence/TopoBench/pull/416}{HMC} & T2 $\cdot$ TNN & topolab & 0.4596 \\
23 & \href{https://github.com/geometric-intelligence/TopoBench/pull/369}{CoGNN} & T1 $\cdot$ GNN & TopoCoop & 0.4585 \\
24 & \href{https://github.com/geometric-intelligence/TopoBench/pull/384}{HeroFilter} & T1 $\cdot$ GNN & TripleA & 0.4574 \\
25 & \href{https://github.com/geometric-intelligence/TopoBench/pull/378}{OGFormer} & T1 $\cdot$ GNN & TripleA & 0.4574 \\
26 & \href{https://github.com/geometric-intelligence/TopoBench/pull/366}{EHNN} & T2 $\cdot$ TNN & LoopCounter & 0.4566 \\
27 & \href{https://github.com/geometric-intelligence/TopoBench/pull/405}{CTNN} & T2 $\cdot$ TNN & howyadoin & 0.4561 \\
28 & \href{https://github.com/geometric-intelligence/TopoBench/pull/365}{GSAN} & T2 $\cdot$ TNN & LoopCounter & 0.4556 \\
29 & \href{https://github.com/geometric-intelligence/TopoBench/pull/345}{Polynormer} & T1 $\cdot$ GNN & SweetLesson & 0.4550 \\
30 & \href{https://github.com/geometric-intelligence/TopoBench/pull/362}{ETNN-LapPE} & T2 $\cdot$ TNN & E(n)igma & 0.4545 \\
31 & \href{https://github.com/geometric-intelligence/TopoBench/pull/391}{ETNN-CoordinatePolicy (structural-LapPE)} & T2 $\cdot$ TNN & E(n)igma & 0.4543 \\
32 & \href{https://github.com/geometric-intelligence/TopoBench/pull/320}{ETNN} & T2 $\cdot$ TNN & E(n)igma & 0.4534 \\
33 & \href{https://github.com/geometric-intelligence/TopoBench/pull/355}{GATE} & T1 $\cdot$ GNN & PushparajD & 0.4457 \\
34 & \href{https://github.com/geometric-intelligence/TopoBench/pull/334}{GFT} & T1 $\cdot$ GNN & TJPaik & 0.4451 \\
35 & \href{https://github.com/geometric-intelligence/TopoBench/pull/417}{WHNN} & T2 $\cdot$ TNN & Remy & 0.4435 \\
36 & \href{https://github.com/geometric-intelligence/TopoBench/pull/350}{DirSNN (Memory-Safe, Harmonic-Aware)} & T2 $\cdot$ TNN & Grold & 0.4414 \\
37 & \href{https://github.com/geometric-intelligence/TopoBench/pull/421}{HyMN} & T1 $\cdot$ GNN & weeyev & 0.4408 \\
38 & \href{https://github.com/geometric-intelligence/TopoBench/pull/371}{Cellular Transformer} & T2 $\cdot$ TNN & TopoCoop & 0.4400 \\
39 & \href{https://github.com/geometric-intelligence/TopoBench/pull/402}{HOD-GNN} & T1 $\cdot$ GNN & SweetLesson & 0.4395 \\
40 & \href{https://github.com/geometric-intelligence/TopoBench/pull/393}{MPSN} & T2 $\cdot$ TNN & Francesco Orsi & 0.4387 \\
41 & \href{https://github.com/geometric-intelligence/TopoBench/pull/359}{SGFormer} & T1 $\cdot$ GNN & AdamGohain & 0.4375 \\
42 & \href{https://github.com/geometric-intelligence/TopoBench/pull/414}{PolynomialFilterTNN (SNN)} & T2 $\cdot$ TNN & r2 & 0.4368 \\
43 & \href{https://github.com/geometric-intelligence/TopoBench/pull/392}{COSIMO} & T2 $\cdot$ TNN & DeveshKaushal & 0.4333 \\
44 & \href{https://github.com/geometric-intelligence/TopoBench/pull/367}{PhenomNN} & T2 $\cdot$ TNN & LoopCounter & 0.4308 \\
45 & \href{https://github.com/geometric-intelligence/TopoBench/pull/411}{GAMMA} & T1 $\cdot$ GNN & Raj Patil & 0.4257 \\
46 & \href{https://github.com/geometric-intelligence/TopoBench/pull/396}{GAUGE (Riemannian Graph Foundation Model)} & T1 $\cdot$ GNN & Lettlini & 0.4219 \\
47 & \href{https://github.com/geometric-intelligence/TopoBench/pull/332}{Neural Sheaf Diffusion (NSD)} & T1 $\cdot$ GNN & TJPaik & 0.4113 \\
48 & \href{https://github.com/geometric-intelligence/TopoBench/pull/387}{Bi-SCNN} & T2 $\cdot$ TNN & MISN\_AY & 0.4057 \\
49 & \href{https://github.com/geometric-intelligence/TopoBench/pull/399}{AdvDIFFormer} & T1 $\cdot$ GNN & LangDiff & 0.4023 \\
50 & \href{https://github.com/geometric-intelligence/TopoBench/pull/394}{AirTNN} & T2 $\cdot$ TNN & Francesco Orsi & 0.4007 \\
51 & \href{https://github.com/geometric-intelligence/TopoBench/pull/420}{GREAD} & T1 $\cdot$ GNN & LangDiff & 0.3937 \\
52 & \href{https://github.com/geometric-intelligence/TopoBench/pull/322}{SheafAN} & T1 $\cdot$ GNN & howyadoin & 0.3887 \\
53 & \href{https://github.com/geometric-intelligence/TopoBench/pull/326}{Neural Sheaf Propagation} & T1 $\cdot$ GNN & howyadoin & 0.3852 \\
54 & \href{https://github.com/geometric-intelligence/TopoBench/pull/413}{HyperGCN} & T2 $\cdot$ TNN & rayquaza & 0.3847 \\
55 & \href{https://github.com/geometric-intelligence/TopoBench/pull/328}{Directed Simplicial Neural Networks (Dir-SNN)} & T2 $\cdot$ TNN & TJPaik & 0.3821 \\
56 & \href{https://github.com/geometric-intelligence/TopoBench/pull/373}{Neural Sheaf Propagation + Holonomy Measurement} & T1 $\cdot$ GNN & Sheafu & 0.3817 \\
57 & \href{https://github.com/geometric-intelligence/TopoBench/pull/388}{FAGCN} & T1 $\cdot$ GNN & MISN\_AY & 0.3813 \\
58 & \href{https://github.com/geometric-intelligence/TopoBench/pull/397}{SMCN} & T2 $\cdot$ TNN & CtrlAltMe & 0.3697 \\
59 & \href{https://github.com/geometric-intelligence/TopoBench/pull/372}{TopoU-Net} & T2 $\cdot$ TNN & TripleA & 0.3666 \\
60 & \href{https://github.com/geometric-intelligence/TopoBench/pull/361}{LINKX} & T1 $\cdot$ GNN & Perelyn Research & 0.3628 \\
61 & \href{https://github.com/geometric-intelligence/TopoBench/pull/329}{AirTNN} & T2 $\cdot$ TNN & r2 & 0.3594 \\
62 & \href{https://github.com/geometric-intelligence/TopoBench/pull/412}{DSNN} & T1 $\cdot$ GNN & howyadoin & 0.3081 \\*
63 & \href{https://github.com/geometric-intelligence/TopoBench/pull/333}{TopNets} & T2 $\cdot$ TNN & TJPaik & 0.2970 \\*
64 & \href{https://github.com/geometric-intelligence/TopoBench/pull/374}{Identity Sheaf Network (ISN)} & T1 $\cdot$ GNN & Sheafu & 0.2755 \\*
65 & \href{https://github.com/geometric-intelligence/TopoBench/pull/407}{GSN} & T1 $\cdot$ GNN & Kajolya & 0.0795 \\
\end{longtable}}

\clearpage
{\footnotesize
\setlength{\LTcapwidth}{\textwidth}
\begin{longtable}{@{}r >{\raggedright\arraybackslash}p{0.40\textwidth} l >{\raggedright\arraybackslash}p{0.20\textwidth} r@{}}
\toprule
\textbf{\#} & \textbf{Model} & \textbf{Track} & \textbf{Team} & \textbf{Score $\downarrow$} \\
\midrule
\endfirsthead
\toprule
\textbf{\#} & \textbf{Model} & \textbf{Track} & \textbf{Team} & \textbf{Score $\downarrow$} \\
\midrule
\endhead
\bottomrule
\caption{Triangle-counting leaderboard for the 65-entry analysis cohort (MSE per triangle, lower is better; Section~\ref{sec:results}). Ranks use full-precision means.}\label{tab:lb_tc}\\
\endlastfoot
1 & \href{https://github.com/geometric-intelligence/TopoBench/pull/418}{CTNN (copresheaf, gated)} & T2 $\cdot$ TNN & CoPreSheafers & 0.0024 \\
2 & \href{https://github.com/geometric-intelligence/TopoBench/pull/410}{SheafTSP} & T2 $\cdot$ TNN & Fernando Espinosa & 0.0108 \\
3 & \href{https://github.com/geometric-intelligence/TopoBench/pull/376}{MPSN} & T2 $\cdot$ TNN & GiggioIlGriggio & 0.0287 \\
4 & \href{https://github.com/geometric-intelligence/TopoBench/pull/393}{MPSN} & T2 $\cdot$ TNN & Francesco Orsi & 0.0333 \\
5 & \href{https://github.com/geometric-intelligence/TopoBench/pull/353}{TopoPolynormer} & T1 $\cdot$ GNN & SweetLesson & 0.0451 \\
6 & \href{https://github.com/geometric-intelligence/TopoBench/pull/392}{COSIMO} & T2 $\cdot$ TNN & DeveshKaushal & 0.0504 \\
7 & \href{https://github.com/geometric-intelligence/TopoBench/pull/419}{ESC-GNN} & T1 $\cdot$ GNN & kxdenz & 0.0552 \\
8 & \href{https://github.com/geometric-intelligence/TopoBench/pull/407}{GSN} & T1 $\cdot$ GNN & Kajolya & 0.0665 \\
9 & \href{https://github.com/geometric-intelligence/TopoBench/pull/370}{Graph Inductive Bias Transformer (GRIT)} & T1 $\cdot$ GNN & TripleA & 0.0693 \\
10 & \href{https://github.com/geometric-intelligence/TopoBench/pull/402}{HOD-GNN} & T1 $\cdot$ GNN & SweetLesson & 0.0802 \\
11 & \href{https://github.com/geometric-intelligence/TopoBench/pull/362}{ETNN-LapPE} & T2 $\cdot$ TNN & E(n)igma & 0.1147 \\
12 & \href{https://github.com/geometric-intelligence/TopoBench/pull/333}{TopNets} & T2 $\cdot$ TNN & TJPaik & 0.1192 \\
13 & \href{https://github.com/geometric-intelligence/TopoBench/pull/406}{DiffLift-SMCN} & T2 $\cdot$ TNN & One Ring to Lift Them All & 0.1201 \\
14 & \href{https://github.com/geometric-intelligence/TopoBench/pull/391}{ETNN-CoordinatePolicy (structural-LapPE)} & T2 $\cdot$ TNN & E(n)igma & 0.1211 \\
15 & \href{https://github.com/geometric-intelligence/TopoBench/pull/320}{ETNN} & T2 $\cdot$ TNN & E(n)igma & 0.1213 \\
16 & \href{https://github.com/geometric-intelligence/TopoBench/pull/404}{$r$-$\ell$MPNN} & T1 $\cdot$ GNN & Oversmooth operators & 0.1382 \\
17 & \href{https://github.com/geometric-intelligence/TopoBench/pull/421}{HyMN} & T1 $\cdot$ GNN & weeyev & 0.1497 \\
18 & \href{https://github.com/geometric-intelligence/TopoBench/pull/387}{Bi-SCNN} & T2 $\cdot$ TNN & MISN\_AY & 0.1498 \\
19 & \href{https://github.com/geometric-intelligence/TopoBench/pull/350}{DirSNN (Memory-Safe, Harmonic-Aware)} & T2 $\cdot$ TNN & Grold & 0.1543 \\
20 & \href{https://github.com/geometric-intelligence/TopoBench/pull/366}{EHNN} & T2 $\cdot$ TNN & LoopCounter & 0.1672 \\
21 & \href{https://github.com/geometric-intelligence/TopoBench/pull/328}{Directed Simplicial Neural Networks (Dir-SNN)} & T2 $\cdot$ TNN & TJPaik & 0.1716 \\
22 & \href{https://github.com/geometric-intelligence/TopoBench/pull/365}{GSAN} & T2 $\cdot$ TNN & LoopCounter & 0.2570 \\
23 & \href{https://github.com/geometric-intelligence/TopoBench/pull/363}{Cell Isomorphism Network (CIN)} & T2 $\cdot$ TNN & TripleA & 0.2638 \\
24 & \href{https://github.com/geometric-intelligence/TopoBench/pull/398}{DPHGNN} & T2 $\cdot$ TNN & Oversmooth operators & 0.4021 \\
25 & \href{https://github.com/geometric-intelligence/TopoBench/pull/414}{PolynomialFilterTNN (SNN)} & T2 $\cdot$ TNN & r2 & 0.4197 \\
26 & \href{https://github.com/geometric-intelligence/TopoBench/pull/394}{AirTNN} & T2 $\cdot$ TNN & Francesco Orsi & 0.4538 \\
27 & \href{https://github.com/geometric-intelligence/TopoBench/pull/416}{HMC} & T2 $\cdot$ TNN & topolab & 0.4843 \\
28 & \href{https://github.com/geometric-intelligence/TopoBench/pull/334}{GFT} & T1 $\cdot$ GNN & TJPaik & 0.4911 \\
29 & \href{https://github.com/geometric-intelligence/TopoBench/pull/371}{Cellular Transformer} & T2 $\cdot$ TNN & TopoCoop & 0.4982 \\
30 & \href{https://github.com/geometric-intelligence/TopoBench/pull/405}{CTNN} & T2 $\cdot$ TNN & howyadoin & 0.5174 \\
31 & \href{https://github.com/geometric-intelligence/TopoBench/pull/332}{Neural Sheaf Diffusion (NSD)} & T1 $\cdot$ GNN & TJPaik & 0.5378 \\
32 & \href{https://github.com/geometric-intelligence/TopoBench/pull/329}{AirTNN} & T2 $\cdot$ TNN & r2 & 0.5401 \\
33 & \href{https://github.com/geometric-intelligence/TopoBench/pull/417}{WHNN} & T2 $\cdot$ TNN & Remy & 0.5492 \\
34 & \href{https://github.com/geometric-intelligence/TopoBench/pull/374}{Identity Sheaf Network (ISN)} & T1 $\cdot$ GNN & Sheafu & 0.6044 \\
35 & \href{https://github.com/geometric-intelligence/TopoBench/pull/321}{SheafHyperGNN} & T2 $\cdot$ TNN & s/pairwise/ho & 0.6742 \\
36 & \href{https://github.com/geometric-intelligence/TopoBench/pull/364}{HiGCN} & T2 $\cdot$ TNN & LoopCounter & 0.8038 \\
37 & \href{https://github.com/geometric-intelligence/TopoBench/pull/397}{SMCN} & T2 $\cdot$ TNN & CtrlAltMe & 0.8219 \\
38 & \href{https://github.com/geometric-intelligence/TopoBench/pull/326}{Neural Sheaf Propagation} & T1 $\cdot$ GNN & howyadoin & 0.8292 \\
39 & \href{https://github.com/geometric-intelligence/TopoBench/pull/360}{Filter Bank GNN (FiGURe)} & T1 $\cdot$ GNN & r2 & 0.8396 \\
40 & \href{https://github.com/geometric-intelligence/TopoBench/pull/354}{PolyFilter GNN (Favard)} & T1 $\cdot$ GNN & r2 & 0.8425 \\
41 & \href{https://github.com/geometric-intelligence/TopoBench/pull/375}{Deep Neural Sheaf Diffusion (DNSD)} & T1 $\cdot$ GNN & Sheafu & 0.8658 \\
42 & \href{https://github.com/geometric-intelligence/TopoBench/pull/345}{Polynormer} & T1 $\cdot$ GNN & SweetLesson & 0.8721 \\
43 & \href{https://github.com/geometric-intelligence/TopoBench/pull/335}{HiD-Net} & T1 $\cdot$ GNN & LangDiff & 0.8753 \\
44 & \href{https://github.com/geometric-intelligence/TopoBench/pull/396}{GAUGE (Riemannian Graph Foundation Model)} & T1 $\cdot$ GNN & Lettlini & 0.8758 \\
45 & \href{https://github.com/geometric-intelligence/TopoBench/pull/415}{GREAD} & T1 $\cdot$ GNN & topolab & 0.8773 \\
46 & \href{https://github.com/geometric-intelligence/TopoBench/pull/408}{DSHN} & T2 $\cdot$ TNN & howyadoin & 0.9073 \\
47 & \href{https://github.com/geometric-intelligence/TopoBench/pull/412}{DSNN} & T1 $\cdot$ GNN & howyadoin & 0.9099 \\
48 & \href{https://github.com/geometric-intelligence/TopoBench/pull/361}{LINKX} & T1 $\cdot$ GNN & Perelyn Research & 0.9132 \\
49 & \href{https://github.com/geometric-intelligence/TopoBench/pull/373}{Neural Sheaf Propagation + Holonomy Measurement} & T1 $\cdot$ GNN & Sheafu & 0.9158 \\
50 & \href{https://github.com/geometric-intelligence/TopoBench/pull/400}{MTGCN} & T1 $\cdot$ GNN & LangDiff & 0.9160 \\
51 & \href{https://github.com/geometric-intelligence/TopoBench/pull/355}{GATE} & T1 $\cdot$ GNN & PushparajD & 0.9223 \\
52 & \href{https://github.com/geometric-intelligence/TopoBench/pull/367}{PhenomNN} & T2 $\cdot$ TNN & LoopCounter & 0.9243 \\
53 & \href{https://github.com/geometric-intelligence/TopoBench/pull/369}{CoGNN} & T1 $\cdot$ GNN & TopoCoop & 0.9286 \\
54 & \href{https://github.com/geometric-intelligence/TopoBench/pull/413}{HyperGCN} & T2 $\cdot$ TNN & rayquaza & 0.9289 \\
55 & \href{https://github.com/geometric-intelligence/TopoBench/pull/411}{GAMMA} & T1 $\cdot$ GNN & Raj Patil & 0.9301 \\
56 & \href{https://github.com/geometric-intelligence/TopoBench/pull/420}{GREAD} & T1 $\cdot$ GNN & LangDiff & 0.9315 \\
57 & \href{https://github.com/geometric-intelligence/TopoBench/pull/352}{Generalized PageRank GNN} & T1 $\cdot$ GNN & PushparajD & 0.9325 \\
58 & \href{https://github.com/geometric-intelligence/TopoBench/pull/399}{AdvDIFFormer} & T1 $\cdot$ GNN & LangDiff & 0.9342 \\
59 & \href{https://github.com/geometric-intelligence/TopoBench/pull/322}{SheafAN} & T1 $\cdot$ GNN & howyadoin & 0.9345 \\
60 & \href{https://github.com/geometric-intelligence/TopoBench/pull/383}{JacobiConv} & T1 $\cdot$ GNN & MISN\_AY & 0.9378 \\
61 & \href{https://github.com/geometric-intelligence/TopoBench/pull/384}{HeroFilter} & T1 $\cdot$ GNN & TripleA & 0.9470 \\
62 & \href{https://github.com/geometric-intelligence/TopoBench/pull/378}{OGFormer} & T1 $\cdot$ GNN & TripleA & 0.9509 \\
63 & \href{https://github.com/geometric-intelligence/TopoBench/pull/359}{SGFormer} & T1 $\cdot$ GNN & AdamGohain & 0.9658 \\
64 & \href{https://github.com/geometric-intelligence/TopoBench/pull/388}{FAGCN} & T1 $\cdot$ GNN & MISN\_AY & 1.0438 \\
65 & \href{https://github.com/geometric-intelligence/TopoBench/pull/372}{TopoU-Net} & T2 $\cdot$ TNN & TripleA & 1.0560 \\
\end{longtable}}

\end{document}